\documentclass[11pt,a4paper]{article}

\usepackage[utf8]{inputenc}
\usepackage[T1]{fontenc}
\usepackage[a4paper,margin=2.5cm]{geometry}
\usepackage{graphicx}
\usepackage{amsmath,amssymb,amsfonts}
\usepackage{amsthm}
\usepackage{appendix}
\usepackage[font=small]{caption}
\usepackage[font=small,justification=centering]{subfig}
\usepackage{url}
\usepackage[round,authoryear]{natbib}
\usepackage[hidelinks]{hyperref}

\theoremstyle{definition}
\newtheorem{asu}{Assumption}
\newtheorem{contribution}{Contribution}

\title{Semantic SLAM in Precision Agriculture using Bayesian Inference}

\author{
  Ruben Beumer\thanks{Corresponding author: \texttt{r.m.beumer@tue.nl}}, Sander Doodeman, Ren\'e van de Molengraft, Duarte Antunes \\[1ex]
  \normalsize Department of Mechanical Engineering, Eindhoven University of Technology \\
  \normalsize P.O. Box 513, 5600 MB Eindhoven, The Netherlands \\
  \normalsize \texttt{\{r.m.beumer, s.doodeman, m.j.g.v.d.molengraft, d.antunes\}@tue.nl}
}

\date{}

\begin{document}

\maketitle

\begin{abstract}
This paper presents a real-time semantic world modeling framework specialized for precision agriculture using autonomous robots.
The framework combines probabilistic mapping of objects and their semantic attributes, updated through Bayesian inference, with a graph-based Simultaneous Localization and Mapping (SLAM) approach implemented using $g^2o$, a general framework for graph optimization.
This integration enables accurate mapping and localization without relying solely on GPS. By leveraging semantic information such as plant type, size, and health, the robot can perform tasks while mapping and localizing itself within a field of crops.
The proposed framework was validated through Gazebo simulations and physical experiments on an indoor field with artificial plants using Boston Dynamics’ robot dog Spot.
A YOLOv8n object detection model was trained to extract object and semantic data from depth camera observations.
These simulations and experiments demonstrate that the system can successfully perform real-time mapping of up to at least 400 plants.
\end{abstract}


\section{Introduction}
\label{sec:intro}
With a declining agricultural workforce and increasing global demand for food, the agricultural sector needs technological innovations. As natural resources become scarcer and sustainability becomes a key aspect, new farming methods must be developed to maximize agricultural yield while minimizing the exhaustion of the soil and damage to the environment, for example by reducing the need for fertilizers. A key element to achieving this goal is precision agriculture.

\subsection{Benefits and challenges of precision agriculture}
The use of technological tools, such as communication systems, data collection methods, and the application of deep learning, enables accurate plant monitoring and treatment, according to the specific needs of specific crops. Numerous studies show that precision agriculture reduces the use of water and fertilizers while increasing yield \citep{Hall1999PrecisionFarming, Cisternas2020PrecisionAgriculture, Gawande2023PrecisionFarming}.

An exemplary farming method that could achieve higher efficiency by using ecological principles is intercropping, where different types of complementary crops are planted together.
There are different types of intercropping techniques.
For temporal intercropping, slow- and fast-growing crops are planted together in the same field in a single season, where one crop type can be harvested earlier than the other.
With row intercropping, rows of different types of plants are planted next to each other and with mixed intercropping, different types of plants are planted simultaneously without distinct row arrangement \citep{Andrews1976MixedIntercropping}.
Different crop types can be matched based on nutrient, sunlight, and water usage, as well as on forming natural pest barriers \citep{Li2020Intercropping, Brooker2015Intercropping}.
In the study by Wang et al., it is claimed that this form of precision agriculture increases the yield by up to 43\% \citep{Wang2023Intercropping}.

\begin{figure}
    \centering
    \includegraphics[width=0.7\linewidth,trim={14cm 26cm 16cm 30cm}, clip]{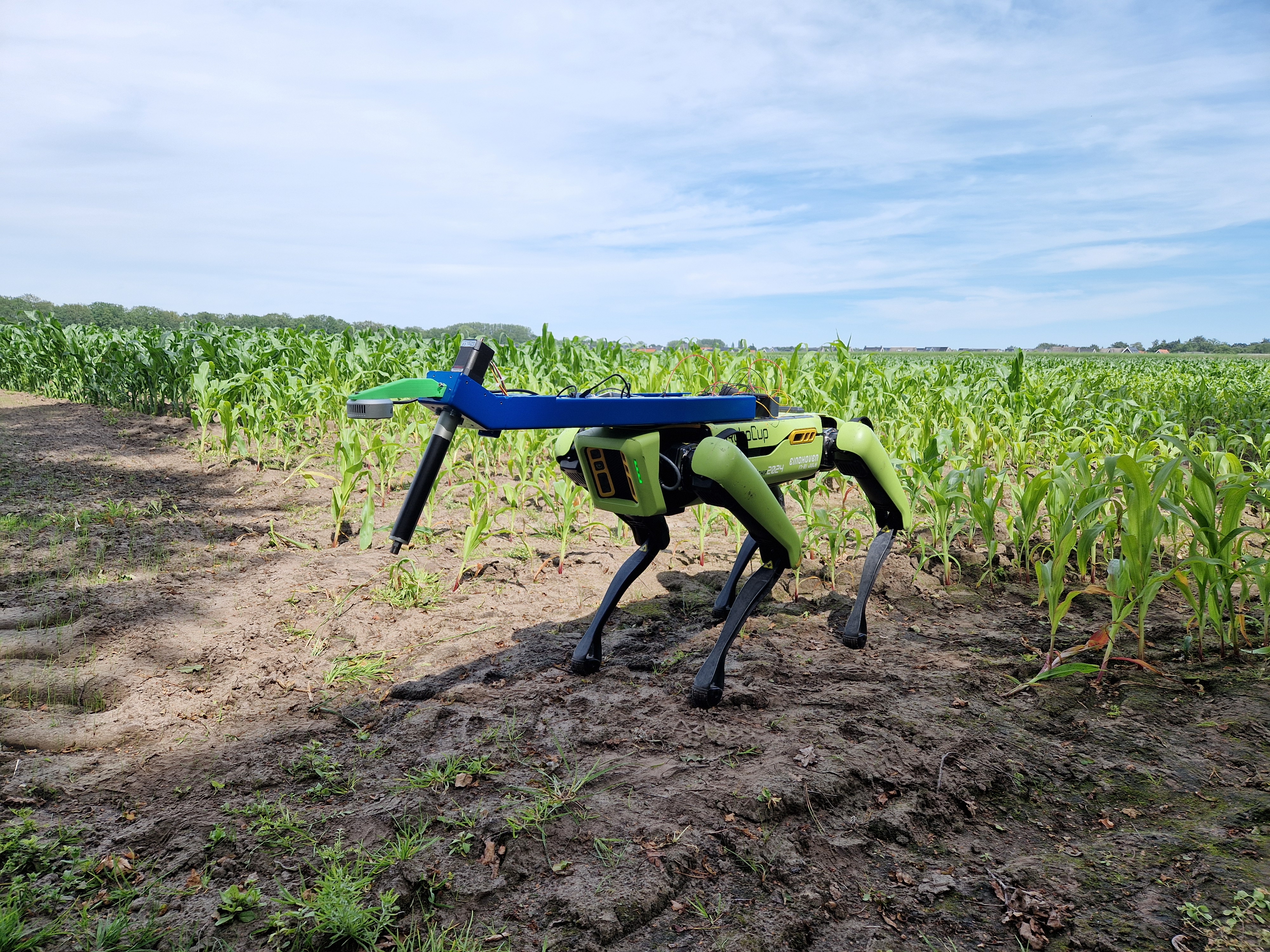}
    \caption{Boston Dynamics Spot with a custom tool for weed removal in a field of maize crops}
    \label{fig:spot-in-field}
\end{figure}

However, precision agriculture comes with big challenges.
In addition to the fact that it requires accurate planning for crop selection and spacing, it is impossible to apply the conventional bulk machines that are currently necessary to achieve the required yield.
This makes precision farming very time-consuming and labor-intensive \citep{Mathews2018Intercropping}, and this is also why intercropping is more prominent in regions with fewer mechanical aids \citep{Huss2022Intercropping}.
For intercropping specifically, challenges arise in the planning of planting, cultivation, fertilization, spraying, and harvesting more than one crop type in a field \citep{Mathews2018Intercropping}.
Although autonomous robots are not as efficient in bulk farming at the moment, they can observe the field on an individual plant level, and also treat the plants on that level, taking care of for example weed and disease control, reducing human labor.
Another very important benefit of using smaller autonomous robots is to reduce the problem of soil compaction. Large agricultural machines put a lot of pressure on the soil, compacting it and reducing the availability of water for the plants, which in turn reduces the yield \citep{Johnson2002Compaction, Chen2011Compaction}.
Moreover, autonomous robots can monitor a crop field in more detail, using specialized sensors, and save this in a database or use this information for decision-making. If the robots are versatile enough, the farmer only has to focus on planning and he/she can monitor the field on a higher level, only focusing on important parts based on the data that the robot gives. For example, robot dogs are versatile and can walk through an agricultural field, as in Fig.~\ref{fig:spot-in-field}. One can attach additional sensors or end effectors to them, either off-the-shelve or custom-made.

Another option is using Unmanned Aerial Vehicles (UAVs) for precision farming purposes, as UAVs can monitor the field from a higher point of view, whereas ground robots cannot always move through a field easily. However, fewer details can be seen from above and the localization and control of UAVs often depend on GPS, which can be unreliable \citep{US2022GPS}. Furthermore, legislation around UAVs can be problematic and they are in general less robust. But, most importantly, ground robots can perform a broader range of tasks than UAVs, as ground robots can carry higher loads, have a longer battery life, and can interact with the plants at crop level.

To perform this range of tasks, such as field monitoring, disease treatment, weed removal, and harvesting, autonomous robots need a sense of understanding the field to be versatile and efficient. Using semantic information is key to achieving that and can be used to program the intentions of a farmer \citep{Achour2022SemanticMapping}. Semantics can be used to provide the farmer with an understandable insight into the field and to interact with the robot.

\subsection{Semantic field mapping}
Research on using semantics in agriculture has so far been limited, especially concerning semantic mapping. Different methods have been proposed to solve the localization problem and the mapping problem, but state-of-the-art mapping techniques, such as topological and semantic mapping, that are applied in different engineering fields, have not yet been developed for agricultural purposes \citep{Aguiar2020AgriculturalMapping}. The challenges that semantic mapping faces concern semantic object detection, choosing the right map representation, communication, and, in the case of multiple robots and/or sensors, sensor fusion. Additionally, applying semantic mapping in agriculture is even more challenging, due to changing lighting conditions and dynamic and less structured environments. For example, crops grow and weeds emerge, so the environment and objects will be different when the robot visits it again after a few days.

Within robotics research, many Simultaneous Localization And Mapping (SLAM) solutions have been provided already. These solutions have varying map representation and data association approaches, and therefore, a lot of inspiration can be drawn from those solutions.

We will now discuss object detection, map representation, and data association problems faced by semantic mapping.

\subsubsection{Semantic object detection}
When it comes to semantic world mapping, it is important to detect and efficiently store specific plants and their properties that the farmer is interested in, such as biomass and ripeness. The plant properties can be complicated to detect using an autonomous robot since sensors and robot movement can be limiting. However, with the rise of machine learning and neural networks, promising methods have been developed in which low-cost sensors, such as vision sensors, can be used to detect the desired properties. An example is YOLOv8, a computer vision model that supports object detection and segmentation \citep{Jocher2023YOLO}. In addition, fluorescence imaging, thermal imaging, and 3D imaging can be used, possibly in combination with deep learning techniques, to detect plant properties, such as diseases \citep{Singh2020PlantDetection, Bhimte2018DiseaseVision}.
Alternatively, nanosensors attached to plants can be used to monitor the plant more accurately and specifically than vision sensors can, enabling more appropriate treatments \citep{Lee2021StaticSensors, Giraldo2019WearableSensor}.
The disadvantage here is that it is difficult to let these fixed sensors cooperate with the autonomous robot. And unless the robot can apply the sensors itself, this would require additional labor by the farmer.

\subsubsection{Map representation}
The detected objects with their semantics have to be saved and kept track of in an efficient way.
A popular means of saving those objects is by using a grid-map representation.
However, this method suffers from a poor scalability due to the use of a fixed resolution, as well as high memory usage, depending on the grid size \citep{Kraetzschmar2004QuadtreeOccupancy}.
This will result in scalability issues, especially for larger agricultural fields.
Another approach is using a point-cloud representation, where the 3D information of objects is saved as part of the point cloud \citep{Weiss2011Pointcloud, Dong2017CropMonitoring}.
However, memory usage is still a problem for this representation.
An alternative that avoids these issues is to use a topological map, in which each object is represented by a node, connected by a measurement edge to a position node of the autonomous robot \citep{Kummerle2011g2o, MurArtal2015ORBSLAM}. The (relative) positions of the nodes contain some uncertainty due to the limited accuracy of the measurements that form the edges between the nodes. Therefore, the same object measured from two different viewpoints will not be detected at the exact same location, and hence an algorithm is required to solve the ambiguity of the nodes' positions. 
This method offers flexibility in the map, and graph optimization algorithms are readily available for fast solving of these nonlinear errors resulting from a graph-based SLAM problem \citep{Kummerle2011g2o}.

\subsubsection{Sensor fusion and data association}
\label{sec:sensor-fusion}
When using multiple robots, multiple sensors, or even multiple measurements regarding the same plants or their attributes, and mapping them in the world map, sensor fusion algorithms are required. More specifically, when measuring the same objects again, data association between object measurements is required to update the right objects in the world map, especially if the robot states (such as position) change and contain a level of uncertainty. One of the simplest solutions to the data association problem is the Global Nearest Neighbor (GNN) method, where new measurements are associated with existing objects by an association matrix, which can be based on the geographic distance, but also based on differences in semantics. Alternatively, a probabilistic approach can be used for the data association, for instance, a Joint Probabilistic Data Association Filter (JPDAF), where an expected value for the state of each object is based on a subset of hypotheses formed by the measurements \citep{BarShalom2009PDAF}. This method uses soft decisions, which is more difficult to handle in combination with a graph-based SLAM approach. Soft decisions retain confidence information about each measurement, allowing the algorithm to weigh uncertain observations appropriately, whereas hard decisions commit to a single outcome and discard uncertainty. With a Multiple Hypothesis Tracking (MHT) approach, if there is uncertainty about measurement associations, multiple hypotheses are regarded and updated individually but simultaneously, rather than concerning only the best hypothesis (as for GNN), or using the weighted average of a combination of hypotheses (as for JPDAF) \citep{Blackman2004MHT}. However, this approach requires significant time and memory resources.
Finally, tracking and fusing objects using neural networks, such as Deep-SORT, is another alternative when detecting objects \citep{Emami2021LearningMOT, Pereira2022SORT}, and can be used in combination with a previously mentioned method.
For instance, the results of a tracking algorithm could be used within the association matrix of the GNN method.
Especially when using vision-based detections, neural networks can easily be applied to track objects. However, if the objects disappear from and reappear in the measurable space, the tracking can be lost, in particular in dynamic environments.
The GNN method is preferred for this work due to its simplicity and high effectiveness in the studied cases. This implies that the framework relies on hard decision-making, for which the associated limitations are not expected to be problematic in fields with sufficient spacing between plants.

\subsection{Active inference}
\label{sec:active-inference}
Once a map with plants and their semantic properties has been created, this information can be used to give the farmer more insight in the field, or for the autonomous robot to make decisions.
The autonomous robot could water the plants or treat diseases based on the gained information.
Next to that, there might be parts of the field that are not measured well enough, and the autonomous robot can be controlled to revisit these parts.
Active inference can be applied to gain more knowledge about the parts of the field that are still uncertain \citep{Friston2009ActiveInference, Catal2020ActiveInference, Friston2012ActiveInference}.
For active inference, the free energy principle can be used, where the free energy represents the variational upper bound of `surprise' in the system, which one generally tries to minimize \citep{Friston2006FreeEnergyPrinciple, Kawahara2021FEP}.
An estimate of the free energy can be calculated for the whole field, and the parts with the highest free energy are most interesting to revisit.
In this work, we introduce an interest heuristic based on the free energy principle for supervisory control purposes.

\subsection{Contributions}
\label{sec:contributions}
This work proposes an agricultural semantic SLAM method, designed for precision agriculture purposes, such as intercropping. This has, to the best of our knowledge, not been presented in literature before. The method developed consists of the following contributions.

\begin{contribution} \label{contribution:main}
    The development of a world model for a probabilistic agricultural semantic world mapping framework that can be updated in real-time, by multiple sensors or robots, exploiting prior knowledge about the structure of the considered field, and broad enough to be used for decision-making applications. This world model is a map with the semantic information of the relevant objects within the world.
\end{contribution}
\begin{contribution} \label{contribution:inference}
    An estimation/inference algorithm to update the semantics in the developed world modeling framework.
\end{contribution}
\begin{contribution} \label{contribution:closedloop}
    A heuristic function meant to be used for active inference, indicating the distribution of uncertainty of information in the field. Such heuristic can for example be used in an active inference algorithm to guide the motion of the robot based on the expected information gain, increasing the efficiency of the robot.
\end{contribution}

The proposed framework that is developed for the above-defined problem uses a probabilistic approach to keep track of the semantic properties of objects, using Bayesian inference to update discrete semantic properties and a Kalman filter to update continuous semantic properties in real time. The localization of the objects and autonomous robot is coped with using the $g^2o$ graph-based SLAM optimization toolbox \citep{Kummerle2011g2o}, including methods to exploit prior knowledge of an agricultural field, such as the existence of rows and the distance between plants or rows. The proposed framework is open to a multi-sensor, and thus multi-robot application, by utilizing a centralized Global Nearest Neighbor data association method. Naturally, a multi-robot application would also require communication between the robots.

The proposed framework can be used in several applications. For instance, for autonomous and precise plant irrigation, where each plant could be given the right amount of water based on measurements. Similarly, this work can be applied to the application of pesticides. Another possible usecase is determining the maturity of crops in a mixed intercropped field, where a robot can, for example with the use of a robot arm, harvest ripe crops autonomously. Furthermore, the framework includes a clear visualization that farmers could use to base decisions on.

\subsection{Organization of this paper}
The remainder of this paper is structured as follows.
In Section~\ref{sec:methodology}, we introduce the semantic world modeling framework that is proposed to map objects and their semantics in a probabilistic way.
The results obtained using this world map are elaborated in Section~\ref{sec:results}, consisting of a simulation part and an experimental part.
Finally, the conclusion and discussion in Section~\ref{sec:conclusion} reflect on the contributions and assumptions.

\section{Proposed semantic SLAM framework}
\label{sec:methodology}
\begin{figure}
    \centering
    \includegraphics[width=0.8\linewidth,trim={6mm 6mm 6mm 6mm},clip]{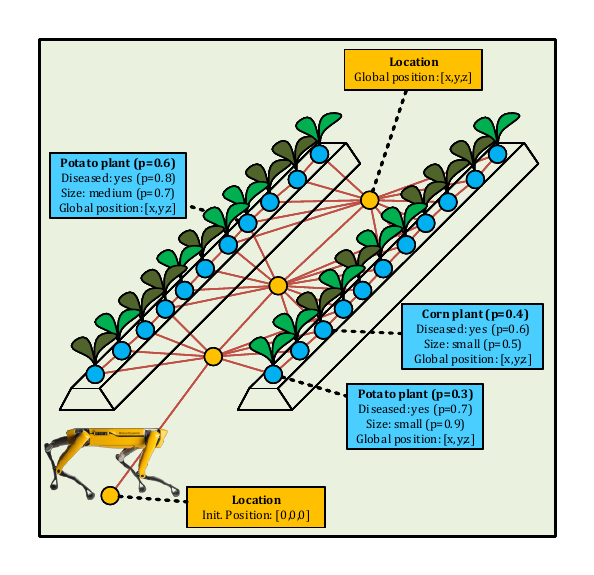}
    \caption{Visual representation of the proposed framework}
    \label{fig:framework}
\end{figure}

In this section, we discuss the developed semantic world mapping framework. In Fig.~\ref{fig:framework}, the idea behind the framework is visualized, where an autonomous robot, Boston Dynamics Spot, is walking along the rows of intercropped plants within a field. The robot provides odometry measurements that connect the robot location nodes, and from these nodes, object (plant) measurements are performed, including different semantic properties. Those semantic properties can either have discrete or continuous values and are updated by Bayesian inference and a Kalman filter, respectively, while the robot is exploring the field. For the localization of the robot and position estimation of the objects, the $g^2o$ graph-based SLAM solution is implemented \citep{Kummerle2011g2o}. This chapter will elaborate on the specific setup of this framework.

\begin{figure*}
    \centering
    \includegraphics[width=\linewidth]{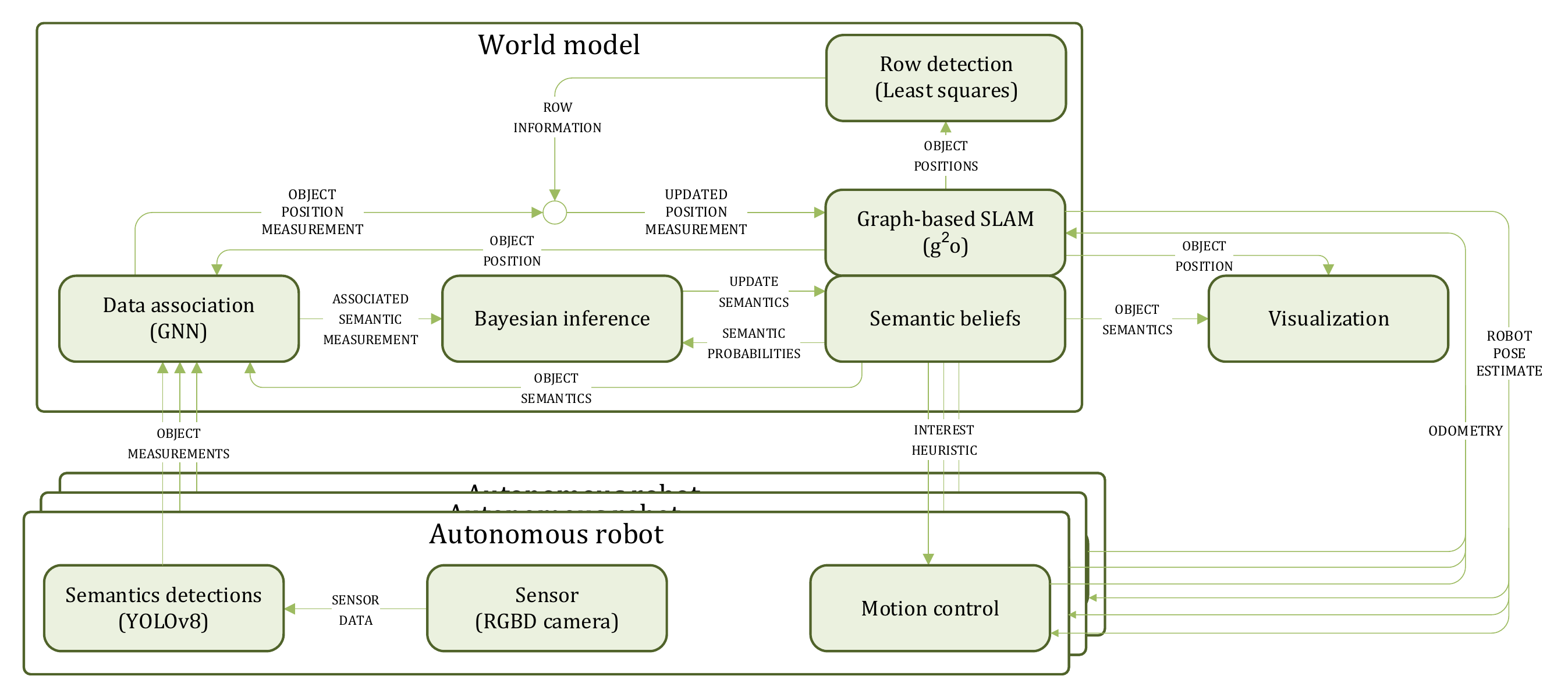}
    \caption{Software flow of the proposed framework}
    \label{fig:codeflow}
\end{figure*}

\begin{figure}
    \centering
    \includegraphics[width=.75\linewidth,trim={0cm, 0.8cm, 0cm, 0cm},clip]{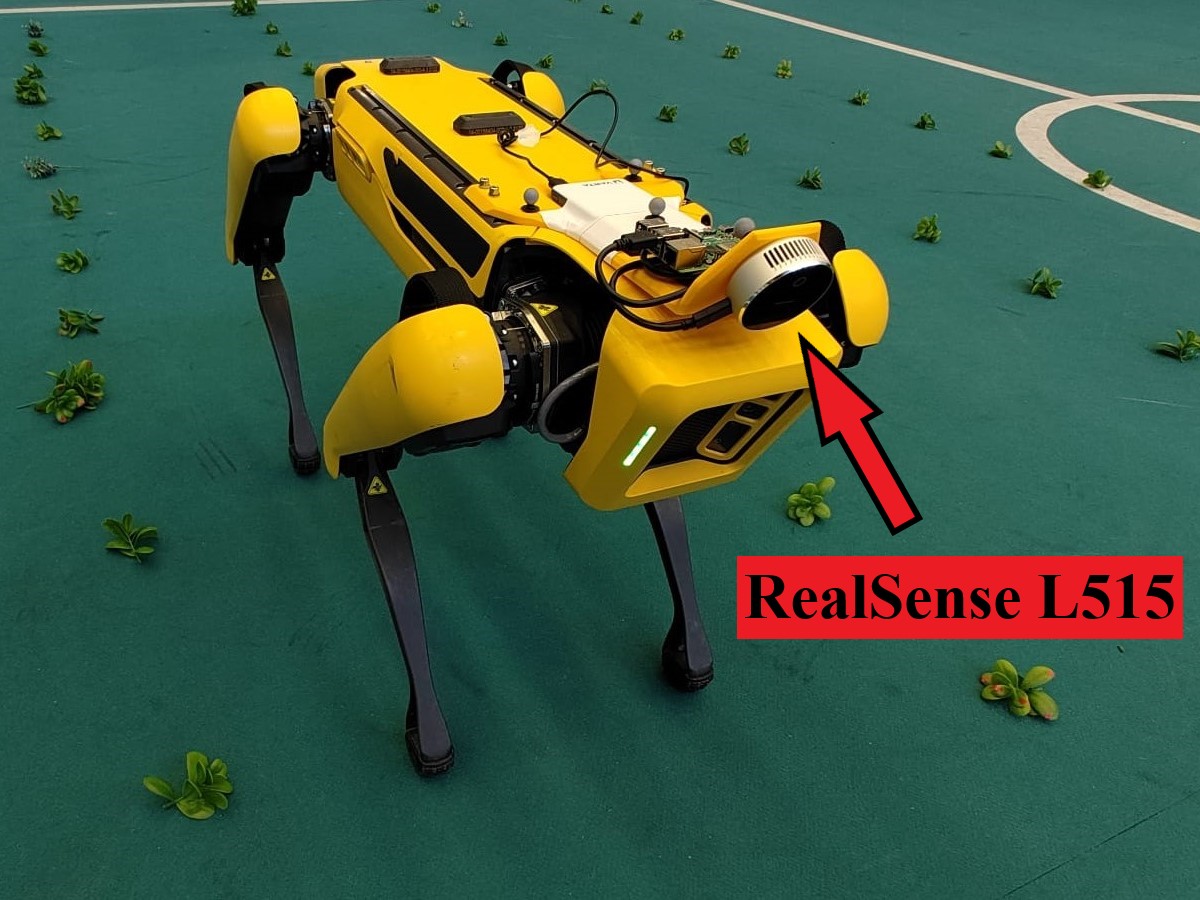}
    \caption{Spot with an L515 RealSense camera on the indoor field with artificial plants}
    \label{fig:spot-in-arena}
\end{figure}

An overview of the software flow is shown in Fig.~\ref{fig:codeflow}. A multi-robot setup is visualized, providing semantic object measurements and odometry measurements to the world modeling framework. The framework has been tested both in a Gazebo simulator with simple plant models and a RealSense RGBD plugin, as well as on an indoor (robot soccer) field with artificial plants, detected by a RealSense L515 LiDAR camera mounted on top of Boston Dynamics Spot, see Fig.~\ref{fig:spot-in-arena}.
The robot has five depth cameras built in. The software by Boston Dynamics for the odometry measurements, used within the proposed SLAM framework, uses these cameras for visual odometry, on top of kinematic odometry. As these cameras do not provide color images, making them less suitable for plant detection, we added the RealSense L515 camera.

\subsection{Preliminaries} \label{sec:preliminaries}
We consider objects with $N$ semantic properties. Each semantic property can be represented by either discrete probability distributions or continuous variables with variances.
Examples of discrete variables used in agriculture are plant types and the presence of diseases.
For continuous variables, plant length and temperature are typical examples found in precision agriculture applications.
For both the discrete and continuous variables, we introduce general notation and entropy terms. The latter are used to indicate the level of uncertainty of the properties of the objects and form a basis for the heuristic function to determine which objects are most interesting to visit, as further explained in Section~\ref{sec:heuristic}.

\subsubsection{Discrete variables}
Each discrete semantic property $i\in\mathcal{X}^d\subseteq\{1, 2, ..., N\}$, has $n_i$ possible values defined by $\mathcal{S}_i = \{s_{i,1}, s_{i,2}, ..., s_{i,n_i}\}$, for instance $\texttt{size} = \{\texttt{small},~\texttt{medium},~\texttt{big}\}$. 
For these variables, a probability distribution for possible values is considered, written as $P(x_i=s_{i,j})~\forall~j\in\{1, 2, ..., n_i\}$, or $P(x_i)$ in short.

We consider a probabilistic approach for object mapping and therefore introduce some entropy terms that are used to quantify probability distributions for variable $x_i$. The Kullback-Leibler divergence (also known as relative entropy, which is used in the free energy principle discussed in Section~\ref{sec:active-inference}) that will be used for discrete variables is given by
\begin{equation*}
{D_\mathrm{KL}(P||Q)}_i = \sum_{s\in\mathcal{S}_i} P(x_i=s) \ln \left( \frac{P(x_i=s)}{Q(x_i=s)} \right).
\end{equation*}
From this definition, we derive a normalized divergence for a probability distribution $P(x_i)$ by normalizing the Kullback-Leibler convergence, given a uniform distribution $Q(x_i=s) = n_i^{-1}~\forall s \in \mathcal{S}_i$, resulting in
\begin{align*}
D_{\mathrm{KLN},i}(P(x_i))&= \frac{1}{\ln(n_i)}\cdot\sum_{s\in\mathcal{S}_i} P(x_i=s) \cdot \ln (P(x_i=s)\cdot n_i) \\
&= 1+\frac{1}{\ln(n_i)}\cdot\sum_{s\in\mathcal{S}_i} P(x_i=s) \cdot \ln (P(x_i=s)).
\end{align*}
This expression is similar to the entropy:~$-\sum_{s\in\mathcal{S}_i} P(x_i=s) \cdot \ln (P(x_i=s))$. However, this normalized Kullback-Leibler divergence has some properties that make it convenient to be used in an active inference framework, as it will equal zero if $P(x_i)$ is a uniform distribution, and it will equal one if, for some $m\in\{1, 2, ..., n_i\}$, ${P(x_i=s_{i,m}) = 1},~{P(x_i=s_{i,j})=0} ~{\forall j \in \{1, 2, ..., n_i\}\setminus m}$.

\subsubsection{Continuous variables}
Alternatively, a continuous semantic property $i\in\mathcal{X}^c = \{1, 2, ..., N\}\setminus\mathcal{X}^d$ can have a continuous range of possible values $\mathcal{S}_i = [s_{i,\mathrm{min}},~s_{i,\mathrm{max}}]$ with $s_{i,\mathrm{min}} \in \mathbb{R} \cup \{-\infty\}$ and $s_{i,\mathrm{max}} \in \mathbb{R} \cup \{\infty\}$, for example, $\texttt{size} = [0,~0.5]$.
The continuous value is accompanied by a variance $\sigma_i^2$.
For continuous probability distributions with infinite support (where in theory $s_{i,\min} = -\infty$ and $s_{i,\max} = \infty$, but where finite bounds are chosen in practice to allow entropy normalization as described below), defined by a Gaussian distribution $\mathcal{N}(s_i, \sigma_i)$, the entropy $H_i$ is given by
\[
{H_i(\sigma_i)} = \frac{1}{2} \ln\left( 2 \pi e \sigma_i^2 \right).
\]
A normalized version of this entropy $H_{\mathrm{N},i}$ is given by
\[
H_{\mathrm{N},i}(\sigma_i) = \frac{1}{2} \ln\left( 2 \pi e \left(\frac{\sigma_i}{s_{i,\max} - s_{i,\min}}\right)^2 \right).
\]

\subsection{Assumptions}
\label{sec:assumptions}
Several assumptions are defined to achieve the contributions stated in Section~\ref{sec:contributions}. Within the proposed framework, an object state includes discrete probability distributions or continuous variables with variances for each semantic property, as given in Section~\ref{sec:preliminaries}. These probability distributions will be updated by individual semantic measurements.
An object measurement consists of a set of measured attributes $\mathcal{Y}$, including at least one semantic property measurement $y_i~\!\in\!~\mathcal{S}_i, \forall~i~\!\in\!~\mathcal{Y}$. For discrete variables, each property measurement $y_i = s_{i,m}$ has an associated probability $p_{i,m}$ of being correct, given the object state, defined by the likelihood function $p_{i,m} = P(y_i=s_{i,m} | x_i=s_{i,m})$ that summarizes the observation model, where $x_i$ represents the real object state.
For this semantic object measurement structure, we first define Assumption \ref{asu:measurements}.
\begin{asu} \label{asu:measurements}
    The probability that another semantic value is present than the measured value $m$, is assumed to be uniformly distributed over the remaining semantic properties:
    \begin{equation}
    \label{eq:uniform-others}
        P(y_i=s_{i,m} | x_i=s_{i,j}) = \frac{1 - p_{i,m}}{n_i-1} \hspace{10mm} \text{ for } j = \{1, 2, ..., n_i\} \setminus m.
    \end{equation}
    This is a compact representation of the sensor model, in terms of $n_i$ possible values for each attribute $i$. Note that $p_{i,m}$ is the probability that measurement result $m$ given by a sensor is correct (i.e., it coincides with the actual value of the attribute). We will use sensor information on this probability to estimate these values $p_{i,m}$ for all $i$ online. Since the sensor provides only the estimated attribute and this probability, this is a convenient model, as further explained in Section~\ref{sec:semantic-object-detection}.
    
    For continuous variables, a Gaussian distribution is assumed, and each property measurement $y_i = s_{i,m}$ has an associated variance. Finally, an object measurement always includes an object position estimate $z$.
\end{asu}

Next, we introduce Assumption~\ref{asu:independence} for the dependence between semantic properties and between measurements.
\begin{asu} \label{asu:independence}
    Each semantic property of an object is independent of the other semantic properties of that object. Moreover, there is no direct dependency between each semantic measurement for an object and any previous measurement for the same semantic property of the same object.
\end{asu}
This assumption is made to offer a simplified world modeling framework, and the interdependence of semantics could be left for future work.

Furthermore, it should be noted that the proposed framework can be applied in 3D, for example in agricultural fields with varying terrain heights. In this work, however, we focus on flat environments and therefore represent all objects in two dimensions. This reduces the number of variables to be estimated within the framework and simplifies both the mapping and localization tasks, as well as the experimental evaluation.

\subsection{Semantic object detection}
\label{sec:semantic-object-detection}
The detection of semantic properties and objects can be done in different ways. In this work we use an RGBD camera:  the L515 RealSense camera on top of Spot for the real experiments, and a simulated camera for the Gazebo simulations.
The proposed framework allows for the detection of a variety of semantics for general objects, and for the plants that are considered in this work, the semantic properties include the plant type, plant size, and diseases. Different types of objects can have different distinct semantic properties that are important to consider. For instance, the health of a weed plant may not be of interest to a farmer, but it is crucial for crops. However, it is still useful to have an estimate for all semantic properties of any detected object, particularly for false detections that may be corrected later. For example, if a plant is initially identified as a weed but is later recognized as a crop, earlier health status measurements can still be used. Therefore, each object in the world model has the exact same semantic properties. After all, the type of plant is considered as a semantic property as well.

To detect these semantic objects from the RGBD color image, a trained YOLOv8n model is used, since it is a state-of-the-art detection model regarding speed and detection quality, while it is still easy to implement \citep{Jocher2023YOLO}.
The model was trained with a relatively small data set; this process is elaborated in Appendix~\ref{app:object-detection}.
The YOLOv8n detection includes a probability indication $p_{i,m}$. Although this probability indication might not exactly be the probability that the measurement property value is the actual property value, given the state of the object (as assumed in Assumption~\ref{asu:measurements}), it provides a useful estimate and is therefore considered as such. In a real application of the framework, $p_{i,m}$ as required by Assumption~\ref{asu:measurements} can be based on data.
Note that if the probability indication $p_{i,m}$ is less than $\frac{1}{n_i}$, the measurement is deemed invalid and hence ignored.
The detection also includes an object segmentation mask, and on top of this, a (green) color filter can be used to get a more accurate image mask for the plants.
This has been implemented for the performed simulations, but not for the experiments that are discussed later, since the color of both the field and the plants is green, and because of a changing exposure in the camera.

For each segmented object, the depth array from the camera is used to calculate an object position by averaging the depth points in the object image mask.
This might cause a bias in the depth estimate towards the robot position, but this is solved by coupling a relatively large depth variance to this measurement within the graph-based SLAM framework (as explained in Section~\ref{sec:graph-slam}).
The resulting position covariance is also used within the data association algorithm (which is partly based on positions, see Section~\ref{sec:data-association}) and should therefore still result in appropriate data association and improved position estimation when multiple measurements are made from different angles.
The above solution worked well for small plants that are relatively far apart. For successful data association for larger plants or smaller distances between plants, this bias in the depth estimate might have to be established and corrected.

\subsection{Data association}
\label{sec:data-association}
For the data association, an association matrix $A$ is developed that can be used by an association algorithm, in this case a GNN algorithm, to associate measurements with existing objects. We define the position difference $e_{mn} = z_n - z_m$, where $z_n$ and $z_m$ are the 2D-positions of the measured object $m$ and the existing object $n$ (following from the graph-base SLAM framework, see Section~\ref{sec:graph-slam}) respectively. We then define $d_\mathrm{inter}$ as the minimum distance between objects, included as prior knowledge, such that the entries $a_{mn}$ of the association matrix $A$ are given by
\[ \begin{aligned}
    a_{mn} =& \overbrace{\prod_{i\in \mathcal{Y}}\left(1- {SSS}_{i,m,n}\cdot w_{\mathrm{sem}}\right)}^{\text{semantic-based}} \\& \cdot \underbrace{\min\left({d_\mathrm{inter}}^{-2} \cdot \lVert e_{mn} \rVert^2,~e_{mn}^\top\cdot K_n^{-1} \cdot e_{mn}\right),}_{\text{distance-based}}
\end{aligned} \]
in which the Semantic Similarity Score (SSS) is given by
\[
    {SSS}_{i,m,n} = \left\{ \begin{array}{cl}
        P\left(x_{i,n}=s_{i,m}\right)\cdot p_{i,m} & \hspace{3mm} \text{if $i\in\mathcal{X}^d$ (discrete variables)} \\
        \tanh(SSS_{i,m,n}^c) & \hspace{3mm} \text{if $i\in\mathcal{X}^c$ (continuous variables)},
    \end{array}\right.
\]
where $P\left(x_{i,n}=s_{i,m}\right)$ is the probability that the object's semantic property $i$ has value $s_{i,m}$ based on prior information and previous measurements (see Section~\ref{sec:bayesian-inference}) and $SSS_{i,m,n}^c$ is given by
\[
    SSS_{i,m,n}^c = \dfrac{1}{\sqrt{2\pi\cdot(\sigma_i^2+\sigma_{i,m}^2)}}\cdot e^{\left(-\dfrac{1}{2}\dfrac{(s_i-s_{i,m})^2}{\sigma_i^2+\sigma_{i,m}^2}\right)},
\]
in which $s_{i,m}$ and $\sigma_{i,m}$ are the measured continuous value and variance. Furthermore, $K_n$ is the position covariance matrix of object $n$, given by the $g^2o$ framework, and ${w_{\mathrm{sem}} \in [0,~1]}$ is a general weight that determines how likely it is that a measured object will be associated with an existing object if the semantics are the same.
If the object measurements are of high quality, $w_{\mathrm{sem}}$ can be relatively high, but if false detections occur often, this value can be lowered.
During data association, no associations are made for values greater than $1$ in the association matrix.
The distance-based term is defined such, that when the distance between the measured object $m$ and the associated object $n$ is larger than $d_\mathrm{inter}$, this measured object is considered a new object unless the associated object has a high position covariance (defined by $K_n$), or when some of the semantics of both objects are the same.
The latter increases this maximum association distance up to a maximum distance defined by ${\max(d_\mathrm{inter},~\sqrt{\lambda_{K_n}}) \cdot {w_{\mathrm{sem}}}^{-|\mathcal{Y}|}}$, where $\lambda_{K_n}$ is the minimum eigenvalue of matrix $K_n$.
The SciPy implementation of a linear sum assignment was used as a GNN approach, which is a modified version of the Jonker-Volgenant algorithm with no initialization \citep{Crouse2016GNN}. Although the YOLOv8n model for object detection can provide object tracking, this was not used within the association matrix, as the tracking results were not consistent enough during the performed experiments.

\subsection{Bayesian inference}
\label{sec:bayesian-inference}
The semantic properties considered in the world modeling framework can be discrete or continuous.
For each discrete semantic property $i$ of any object in the world model, a probability distribution $P(x_i)$ is calculated using Bayesian inference after every measurement. For continuous variables, the probability distribution is also updated after every measurement, which means updating the values of these variables and their variances.
In addition, the probabilities can be updated over time according to a time-dependent model.

The advantage of a continuous variable is that it is less memory-consuming, especially if a higher accuracy is required, since only one value and variance have to be updated for each measurement, and it is easier to update this variable if a continuous dynamic model is present for this variable. The advantage of using discrete variables is that the variable and measurement do not have to follow a Gaussian distribution, and some properties, such as the type of plant, can only be represented as discrete variables.

\subsubsection{Discrete variable measurement update}
In this work, the probability distribution for discrete variables is initialized as a uniform distribution when a new object is measured.
In a real application, the initial probability distribution can be based on prior knowledge.
If during mapping a measured object is associated with an object in the world model, the semantic properties of that object will be updated.

For discrete variables, this is done by Bayesian inference, given by
\[ \text{posterior} = \frac{\text{likelihood} \cdot \text{prior}}{\text{evidence}}. \]
For a semantic property measurement $y_i = s_{i,m}$, the update for $P(x_i)$ is given by
\[ \begin{aligned}
P(x_i=s_{i,j} | y_i=s_{i,m})&= \frac{P(y_i=s_{i,m} | x_i=s_{i,j}) \cdot P(x_i=s_{i,j})}{P(y_i=s_{i,m})}\\
&= \frac{P(y_i=s_{i,m} | x_i=s_{i,j}) \cdot P(x_i=s_{i,j})}{\sum_{k=1}^{n_i} P(y_i=s_{i,m} | x_i=s_{i,k}) \cdot P(x_i=s_{i,k})}.
\end{aligned}\]
Here, $P(x_i=s_{i,j})$ is the prior probability for semantic value $s_{i,j}$, based on previous measurements. In our application, $P(y_i=s_{i,m} | x_i=s_{i,m})$ equals $p_{i,m}$. Additionally, according to Assumption \ref{asu:measurements}, all unknown likelihoods are assumed to be equally distributed following \eqref{eq:uniform-others}. However, in a real application, this likelihood probability distribution should be based on measurement data, and could also be affected by factors such as light conditions.

\subsubsection{Discrete variable time update} \label{sec:discrete-time-update}
After each time step, i.e. before each measurement, the probability distribution of a semantic variable can be updated based on a time-dependent model. For instance, the size of plants can increase over time, or it will be more likely that a plant will get a disease.
A semantic variable does not always have to be updated over time.
To illustrate, the type of plant will not change over time. Given a time step $\Delta t$, an update matrix $U$ can be formed, for example, based on a discrete model for the semantic property (as shown in Appendix~\ref{app:time-update}). On top of that, noise $w$ can also be included in this update matrix, such that the probability distribution changes over time:
\[
P\left(x_i(t+\Delta t)=s_{i,j}\right)= U(\Delta t, w) \cdot P\left(x_i(t)=s_{i,j}\right).
\]

\subsubsection{Continuous variable time update}
For continuous semantic property updates, a Gaussian probability distribution is assumed in Assumption~\ref{asu:measurements}. Therefore, as a special case of Bayesian inference, a Kalman filter is used to update the variables. An example of such a variable could be plant size or temperature, which can be approximated as a Gaussian variable as long as the measurement is accurate enough. This distribution should be initialized as a normal distribution, but similar to the discrete probability distribution, the initial distribution can be based on prior knowledge. 

Note that continuous variables can also be discretized, such that Bayesian inference can be used as an approximation for the probability distribution of that variable. This is especially useful when the variable is not Gaussian, but this discretization comes with a memory usage cost, depending on the desired resolution. The position of a world object is not handled as described above, even though it is a continuous variable. Instead, a graph-based SLAM optimization approach was used to update this property.

\subsection{Graph-based SLAM}
\label{sec:graph-slam}
\begin{figure}
    \centering
    \includegraphics[width=0.6\linewidth]{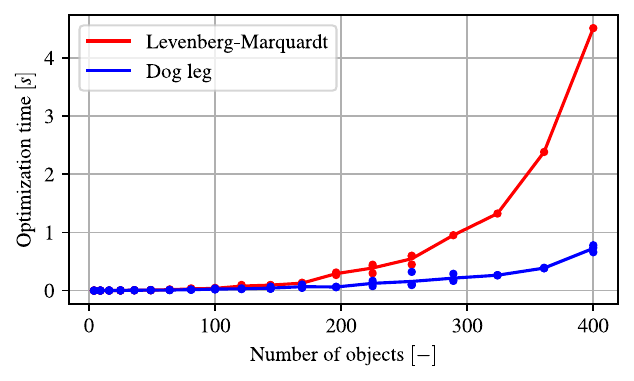}
    \caption{Time scaling of optimization algorithms (Levenberg-Marquardt and Powell's dog leg method) for increasing number of objects.}
    \label{fig:scaling}
\end{figure}

For updating the position estimates of both the autonomous robot and the detected objects, a graph-based SLAM approach is used, by implementing $g^2o$, a simple and efficient tool for graph-based SLAM \citep{Kummerle2011g2o}.
Although the position could also be implemented as a Gaussian variable, or discretized variable within the used framework, $g^2o$ was used, partly because it offers a graph-based representation of the world, including loop closure, to ensure a consistent world map, which also helps to account for odometry noise.
Furthermore, $g^2o$ is easy to implement and relatively fast.
In this section, more features of the $g^2o$ framework are described that show its advantage.
In Fig.~\ref{fig:scaling}, the optimization times for different numbers of objects are given for different optimization methods (the Levenberg–Marquardt algorithm and Powell's dog leg method \citep{Marquardt1963Levenberg, Powell1970dogleg}) with qualitatively similar results. For these optimizations, an HP ZBook Studio G4 was used, with 8 Intel\textsuperscript{\textregistered} Core\textsuperscript{TM} i7-7700HQ CPUs, at $2.80~[GHz]$ and a Quadro M1200 Mobile GPU. With this setup, only when the number of objects is greater than 400, the optimization time can be a problem for real-time applications. In that case, speed improvements must be considered, such as splitting the measurement area into parts, or performing the graph optimization step less often, for instance, when the odometry measurements are of high quality. A difficulty within the $g^2o$ graph-based SLAM framework is that it considers landmarks to be unique, while for an agricultural field, plants can be almost identical. Therefore, data association is an important feature within this framework.

The $g^2o$ implementation consists of nodes belonging to the SE(2) group (for robot locations, including $x$, $y$, and yaw $\theta$) and nodes with Euclidean $(x,~y)$ coordinates (for objects), all of which can be connected by edges. Measured objects are either associated with existing objects or added as a new node to the graph with an edge to the current location.
When associated with existing objects, an additional edge is created between the current location and the existing object. Locations are connected based on odometry measurements from Spot.
Every edge also includes an information matrix, which is the inverse of the covariance matrix. For odometry data, this covariance matrix is a $3 \times 3$ matrix with the covariances for $\Delta x$, $\Delta y$, and $\Delta \theta$ in the local robot frame from the odometry data. Although there might be interdependence between $\Delta x$, $\Delta y$, and $\Delta \theta$, we chose to neglect this influence for the simulations and experiments. For object measurements, the covariance matrix is a $2 \times 2$ matrix, with the position covariance for the $x$- and $y$-position relative to the robot, where the distance to the robot has a higher covariance than the covariance perpendicular to this distance.

\subsubsection{Line intersection}
\begin{figure}
    \centering
    \includegraphics[width=0.85\linewidth]{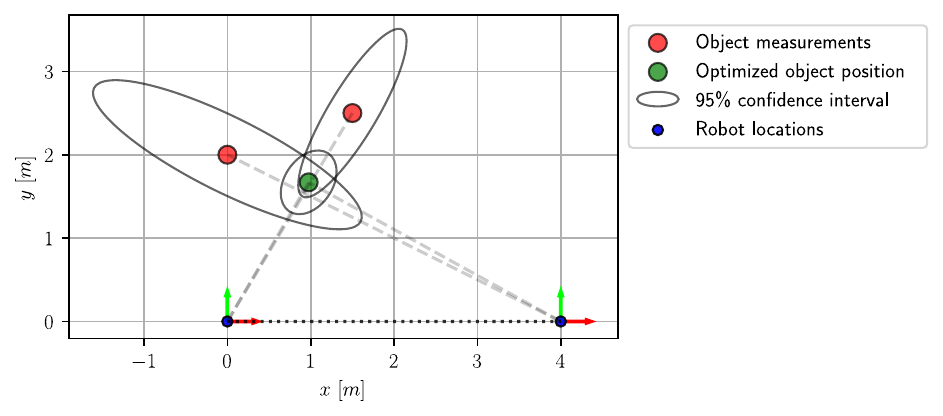}
    \caption{`Line intersection' optimization for two object measurements for the same object}
    \label{fig:line-intersection}
\end{figure}

An advantage of this framework is that two measurements from different robot locations, if correctly associated, can result in an accurate position estimate, even when the depth measurement is uncertain, as long as the direction measurement is certain. This is especially useful when using vision for object and position estimates, since the direction to the object's location from the camera view is relatively certain, while the depth is more difficult to establish. This depth uncertainty can originate from uncertainty from the camera or the image segmentation for the detected object. An example of this is shown in Fig.~\ref{fig:line-intersection}.

\subsubsection{Row information}
\label{sec:row-information}
\begin{figure}
    \centering
    \includegraphics[width=0.9\linewidth]{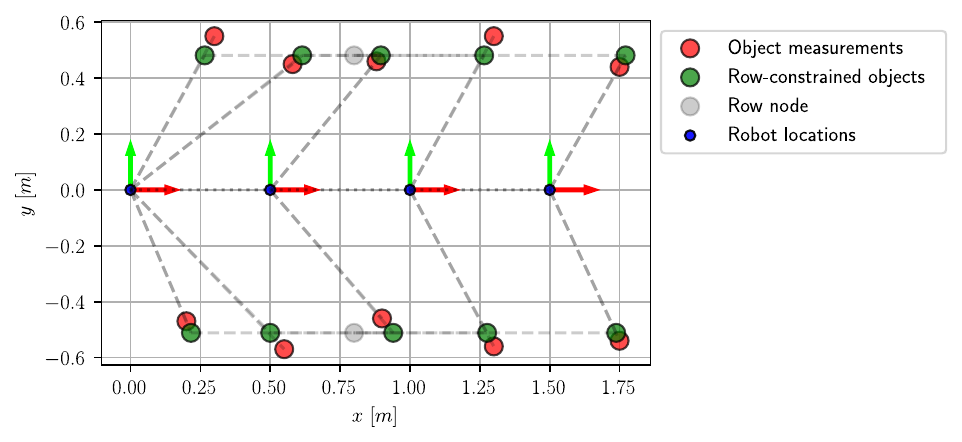}
    \caption{Using prior knowledge about rows by adding a soft constraint between plants and a row}
    \label{fig:row-information}
\end{figure}

Furthermore, prior knowledge about rows within a field can be included and used to improve the position estimate of objects, thus also improving the data association for those objects.
This is done by adding a Euclidean $(x,~y)$ node representing a row, placed at an arbitrary position within that row.
All plants within that row will get a soft constraint to this added node (and thus to the row) using edges with a corresponding information matrix.
In Fig.~\ref{fig:row-information}, an example can be seen, where ten (artificial) object measurements are made, which can get a soft constraint to a row node, and an information matrix. In this case, the objects and row node are connected by an edge with an information matrix that allows little distance along the $y$-direction, while allowing any distance along the $x$-direction. Within the proposed world modeling framework, the row positions can be along any line. The positions of these rows, formed by the angle with respect to the x-axis, and the distance between the first row and the origin, are calculated using a least squares method after $n_{\text{det,est}}$ measurements. After this calculation, soft constraints are added between already existing objects and the rows.

In addition, the distance between rows $d_\mathrm{rows}$ can be softly constrained by edges between the special row nodes.
Here, it is assumed that the distance between the rows is known.
Note that not all plants have to be assigned to a row (see also Appendix~\ref{app:not-in-row}). For example, if the world modeling framework is also used for weed detection, it is less likely that these weed plants are aligned with the crops along a row.

\subsubsection{Handling false detections}
The robot may produce false detections, either by detecting the same object twice within a single view or by incorrectly identifying a feature in the environment as an object. To handle such cases, additional components in the algorithm are implemented to merge objects that are closer together than the minimum inter-plant distance $d_\mathrm{inter}$ and to remove objects that have been detected only a few times and with high uncertainty. This uncertainty is determined with the use of an interest heuristic.

\subsection{Interest heuristic for semantic-based active inference}
\label{sec:heuristic}
For autonomous robots in precision agriculture, navigating the field row-by-row, covering the entire field again every time, is not always the most efficient. For both monitoring and execution tasks related to, for example, weed and disease control, efficiency could benefit from using the constructed map with information based on earlier measurements, possibly combined with prediction models. This map does not only contain information about the current status of the field (e.g., plant growth, weed and disease spreading), but also about the uncertainty of this information, so that plants or entire areas that contain more uncertainty can be visited with priority using an active inference framework.
For this purpose, we calculate an interest heuristic $h$ for each object in the map. This heuristic is a mathematical expression that has the objective of indicating whether it is interesting to revisit objects that the world model is unsure about or that have been measured only a few times. Therefore, it consists of a probability distribution-based term and a term based on the number of measurements for this object:
\begin{equation}
\label{eq:heuristic}
    h = 1 - \underbrace{\frac{n_\mathrm{meas}}{n_\mathrm{meas}+1}\vphantom{\sum_{i=1}^{N}}}_{\substack{\text{number of} \\ \text{measurements}}} \cdot \underbrace{\frac{1}{N}\cdot\sum_{i=1}^N 1-E_{\mathrm{N},i}}_\text{probability distribution},
\end{equation}
in which $n_\mathrm{meas}$ is the number of measurements of the object considered (possibly within a given period) and $E_{\mathrm{N},i}$ is based on entropy and is given by
\[
    E_{\mathrm{N},i} = \left\{ \begin{array}{cl}
        1-D_{\mathrm{KLN},i}(P(x_i)) & \text{ if $i\in\mathcal{X}^d$} \\
        \ell (H_{\mathrm{N},i}(\sigma_i)) & \text{ if $i\in\mathcal{X}^c$}, 
    \end{array} \right.
\]
where $\ell(x)$ is a sigmoid function given by
\[
    \ell(x) = \frac{1}{1+e^{-\alpha(x-\beta)}}.
\]
Throughout this work, the parameter values in this sigmoid function are chosen as $\alpha=2$ and $\beta=0.2$.

The two factors presented in \eqref{eq:heuristic} are used, as the uncertainty about a measured object is more relevant if this object has only been measured a few times, compared to when it is measured relatively often and additional measurements are less likely to improve the certainty about the object. On the other hand, an object that was measured with high confidence is not interesting to revisit, even if it has only been measured a few times.

With the provided setup of the world modeling framework, including the interest heuristic, it is possible to let the robot walk through a field autonomously. Within this project, both for the simulator and experiments, a predefined path is first used to let Boston Dynamics Spot walk around a field of plants. Here, the graph-based SLAM positions are used to make sure Spot walks in between the rows of plants, and not on top of the rows of plants due to odometry and control errors. After the predefined path, the robot could be moved based on the interest heuristic, for instance, by visiting the plants with the highest interest heuristic again.

\section{Results}
\label{sec:results}
This chapter elaborates on the simulations and experiments that were performed using the proposed framework with a (virtual and real) Spot robot dog.
In simulations it is easier to do large-scale and reproducible tests with any number of plant attributes and possible values and a perfectly known ground truth, whereas the experiments with the physical robot, camera and plants, the entire pipeline can be tested in a more realistic setting to provide a proof of concept.

\subsection{Simulations} \label{sec:simulations}
We will first discuss the setup of the simulation, its results, and the feasibility of real-time application. The results will primarily be assessed in terms of accuracy of the SLAM framework, also under the presence of (artificially added) noise, and how computation time scales with the number of plants.

\subsubsection{Simulation setup}
\begin{figure}
    \centering
    \includegraphics[width=0.75\linewidth]{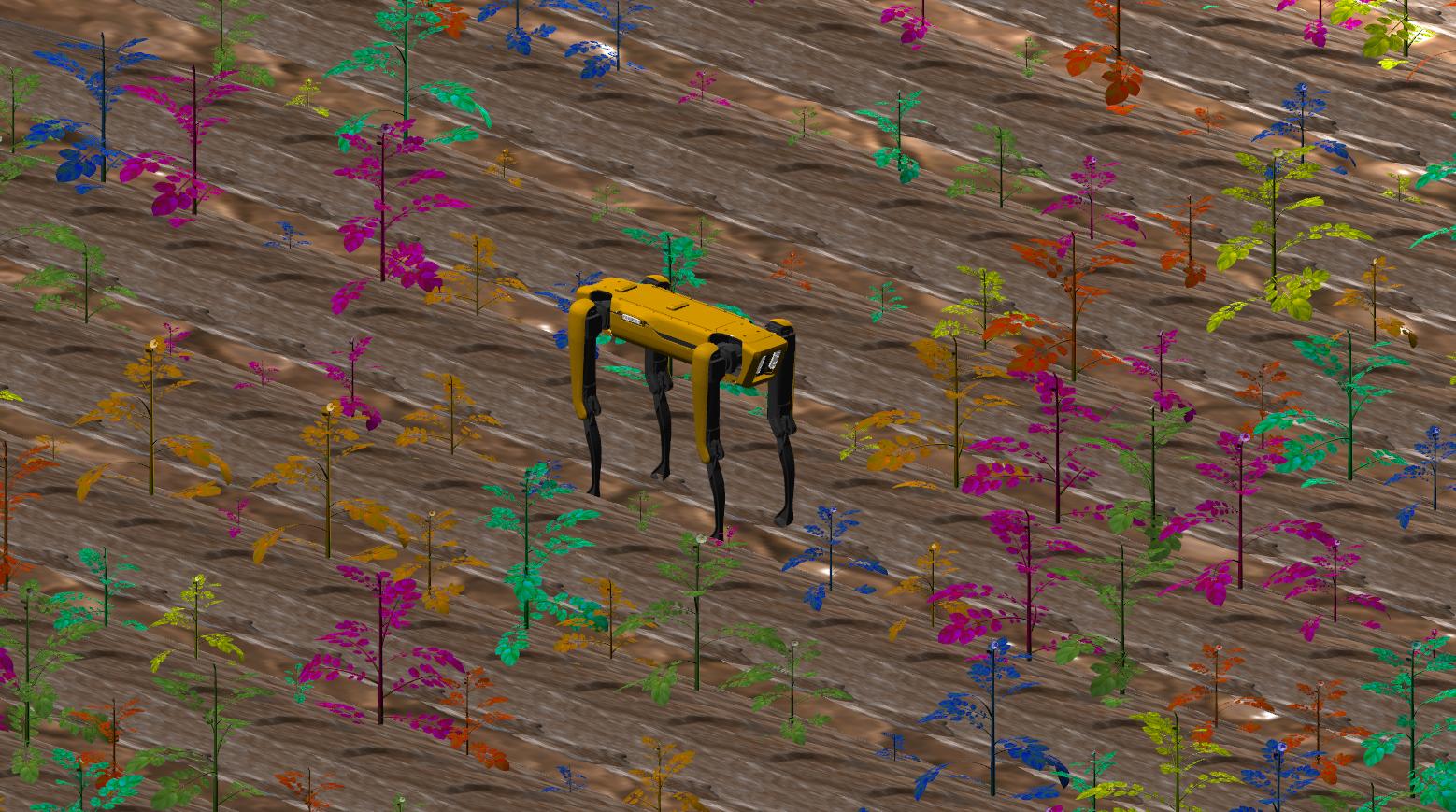}
    \caption{Boston Dynamics Spot in a Gazebo simulator with artificial colored plants}
    \label{fig:spot-in-simulation}
\end{figure}

For testing the framework and demonstrating its benefits, a Gazebo simulation was set up, based on previous work \citep{Devillers2023Master, Narasimhakartheesan2023Master}. It contains a virtual Spot robot dog with two depth cameras in front, in a field with rows of identical plants, which only vary in color, size, and rotation as can be seen in Fig.~\ref{fig:spot-in-simulation}.
The distinct colors represent a property that can be detected by a neural network just as the type of plant in a real field can also be detected by a neural network, where the use of different colors requires less training, also because the image segmentation can then be done by color filtering.
The colors also provide a clear base for the interpretation of the results.

In the simulations, the rows are 1~m apart, and the distance between plants was uniformly distributed between 0.4 and 0.5~m. For data association, $w_{\mathrm{sem}}=0.3$ was chosen, as false detections occurred regularly. The number of object measurements after which the row estimation is performed is $n_{\text{det,est}} = 12$ (see Section~\ref{sec:row-information}) measurements.
A summary of the different plant attributes and their possible values used for the simulations is given in Table~\ref{tab:attributes-sim}.

\begin{table}[h]
    \centering
    \caption{Plant attributes and their possible values used for the simulations}
    \label{tab:attributes-sim}
    \begin{tabular}{l|l|l}
        \textbf{Simulation} & \textbf{Plant type/color} & \textbf{Size} \\ \hline
        Discrete plant size & \texttt{blue}, \texttt{cyan}, \texttt{green}, \texttt{orange}, \texttt{pink}, \texttt{red}, \texttt{yellow} & \texttt{small}, \texttt{medium}, \texttt{big} \\ \hline
        Continuous plant size & \texttt{blue}, \texttt{cyan}, \texttt{green}, \texttt{orange}, \texttt{pink}, \texttt{red}, \texttt{yellow} & $\mathbb{R}$
    \end{tabular}
\end{table}

\subsubsection{Simulation results}
\begin{figure}
    \centering
    \includegraphics[width=0.85\linewidth]{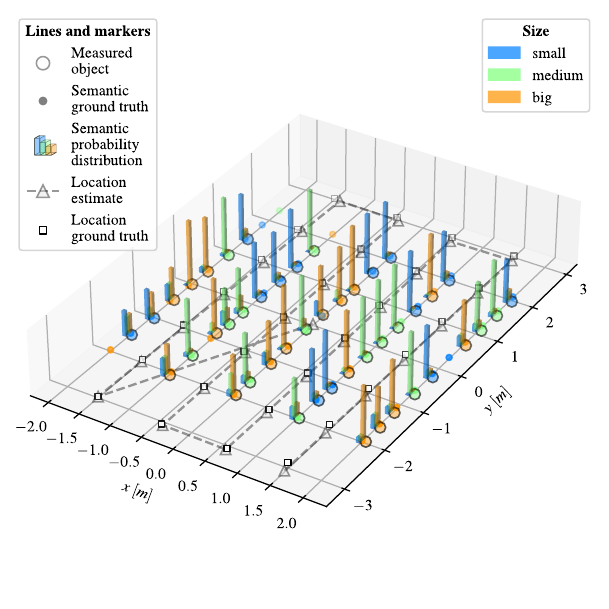}
    \vspace{-1cm}%
    \caption{Simulation example where the size probability distribution is visualized with the use of 3D-histograms}
    \label{fig:barplot}
\end{figure}

\paragraph{Discrete plant size}
In Fig.~\ref{fig:barplot}, an exemplary simulation result is visualized for 50 plants and one surveillance round. Of these 50 plants, the size of 44 plants was estimated correctly, where 45 plants were detected in total. Regarding the plant colors that were also detected with the YOLOv8n network, all colors were detected correctly out of the 43 measured plants.

\begin{figure}
    \centering
    \includegraphics[width=0.85\linewidth]{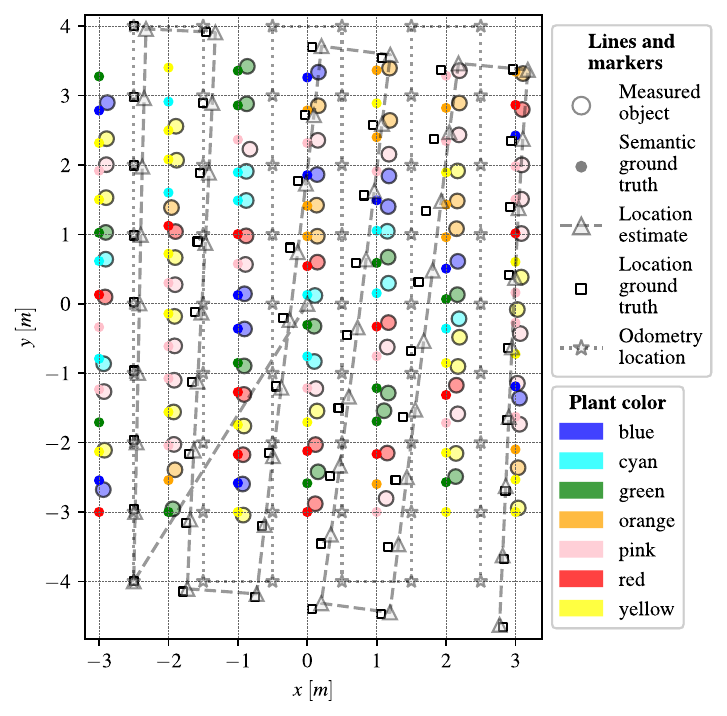}
    \caption{Odometry noise cancellation for a simulation starting at the left bottom $(-2.5, -4)$, and ending at the right bottom of the figure $({\sim}2.5, {\sim}-5)$.}
    \label{fig:odom-noise}
\end{figure}

Simulations in Gazebo where noise was added to the robot odometry input, show that the position estimate using the proposed framework with graph optimization is closer to the actual position (including noise) than the estimate solely based on the odometry, see Fig.~\ref{fig:odom-noise}. Although the end of the simulation shows a somewhat larger error, this can be attributed to the absence of measurements at the end of the row, such that the optimization is done based on less information. The RMS error is 0.503~m without and 0.144~m with graph optimization, demonstrating the benefit of applying it. Simulations in Appendix~\ref{app:simulations} show that when the localization error of the robot is small enough and when at the end of the simulation the robot returns to a place that has been visited at the beginning, the robot correctly associates the plants again, ensuring a consistent world map.

\begin{figure}
    \centering
    \includegraphics[width=0.85\linewidth]{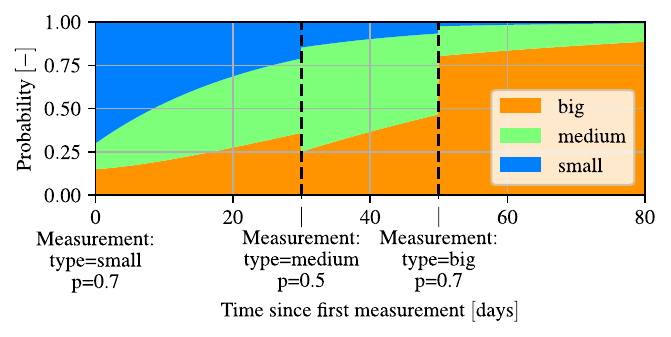}
    \caption{A probability distribution for plant size over time, with Bayesian inference for three measurements of the same plant, including a growth model}
    \label{fig:bayes-over-time}
\end{figure}

In Fig.~\ref{fig:bayes-over-time}, the probability distribution for the three different plant sizes is visualized over time, where a state model over time is used, as described in Section~\ref{sec:discrete-time-update}.
The plant size probability distribution is not only updated by new measurements but also by a growth model.
Plant growth is not included in the Gazebo simulation, and for the experimental setup, no real growing plants are used. Therefore, fictional measurements are considered instead.
In this figure, a plant is considered to grow asymptotically over time, increasing the plant size, using the growth model described in Appendix~\ref{app:time-update}. New measurements update the probability distribution based on the prior probability at that time.

\begin{figure}
    \centering
    \includegraphics[width=0.9\linewidth]{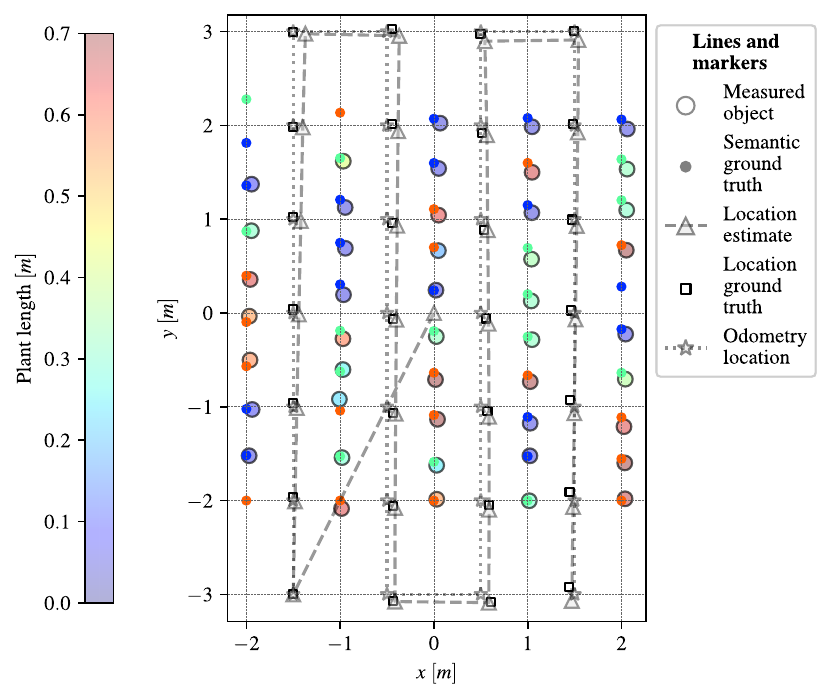}
    \caption{Continuous length estimates with solid circles as ground truth}
    \label{fig:lengths-cont}
\end{figure}

\paragraph{Continuous plant size}
In another simulation, the length of the plant was introduced as a continuous variable. This length was estimated using the bounding boxes in the YOLOv8n detections, in combination with the distance of the plants to the robot, and the variance of the length measurement was scaled quadratically with this distance. This results in Fig.~\ref{fig:lengths-cont}, where three different sizes of plants were present in the simulation, and a root-mean-square error for the estimated plant length is 0.118~m, for plants with lengths between 0.12 and 0.57~m.

\subsubsection{Computation time}
\begin{figure}
    \centering
    \includegraphics[width=.7\linewidth]{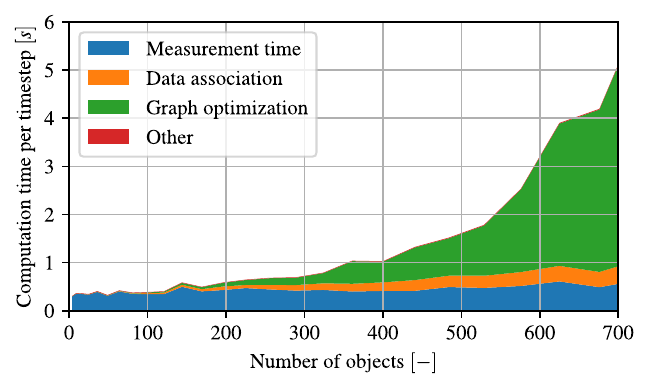}
    \caption{Computation time per time step for different numbers of plants, decomposed in different steps of the proposed framework}
    \label{fig:sim-times}
\end{figure}

The feasibility of real-time application of the proposed framework was also investigated using simulations, as large numbers of plants can easily be simulated. In Fig.~\ref{fig:sim-times}, the computation time per time step is shown, decomposed in the different steps in the proposed framework. Also for these simulations, an HP ZBook Studio G4 was used, with 8 Intel\textsuperscript{\textregistered} Core\textsuperscript{TM} i7-7700HQ CPUs, at 2.80~GHz and a Quadro M1200 Mobile GPU, where the Gazebo simulation and world modeling framework were executed simultaneously in an Ubuntu operating system. It is clear that the feasibility of real-time application is highly dependent on the graph optimization times. Considering that measurements are made approximately once per second, the proposed world modeling framework can run in real time for up to 400 objects. If the framework would be applied to larger fields, as was explained in Section~\ref{sec:graph-slam}, speed improvements could be made by splitting the field into parts, or performing the graph optimization less often in order to apply the proposed framework in real time in any real crop field of any size. The minimum required update frequency depends on the level of drift in odometry data and measurement noise, so that it does not cause incorrect data association, which might cause positioning errors to accumulate.

\subsection{Experiments} \label{sec:experiments}
Also for the experiments with the real robot and (artificial) plants, we will discuss the setup and results.

\subsubsection{Experimental setup}
\begin{figure}
\centering
\subfloat[Small diseased crop (\texttt{A})]{\includegraphics[width=.32\linewidth,trim={0cm, 6cm, 0cm, 12cm},clip]{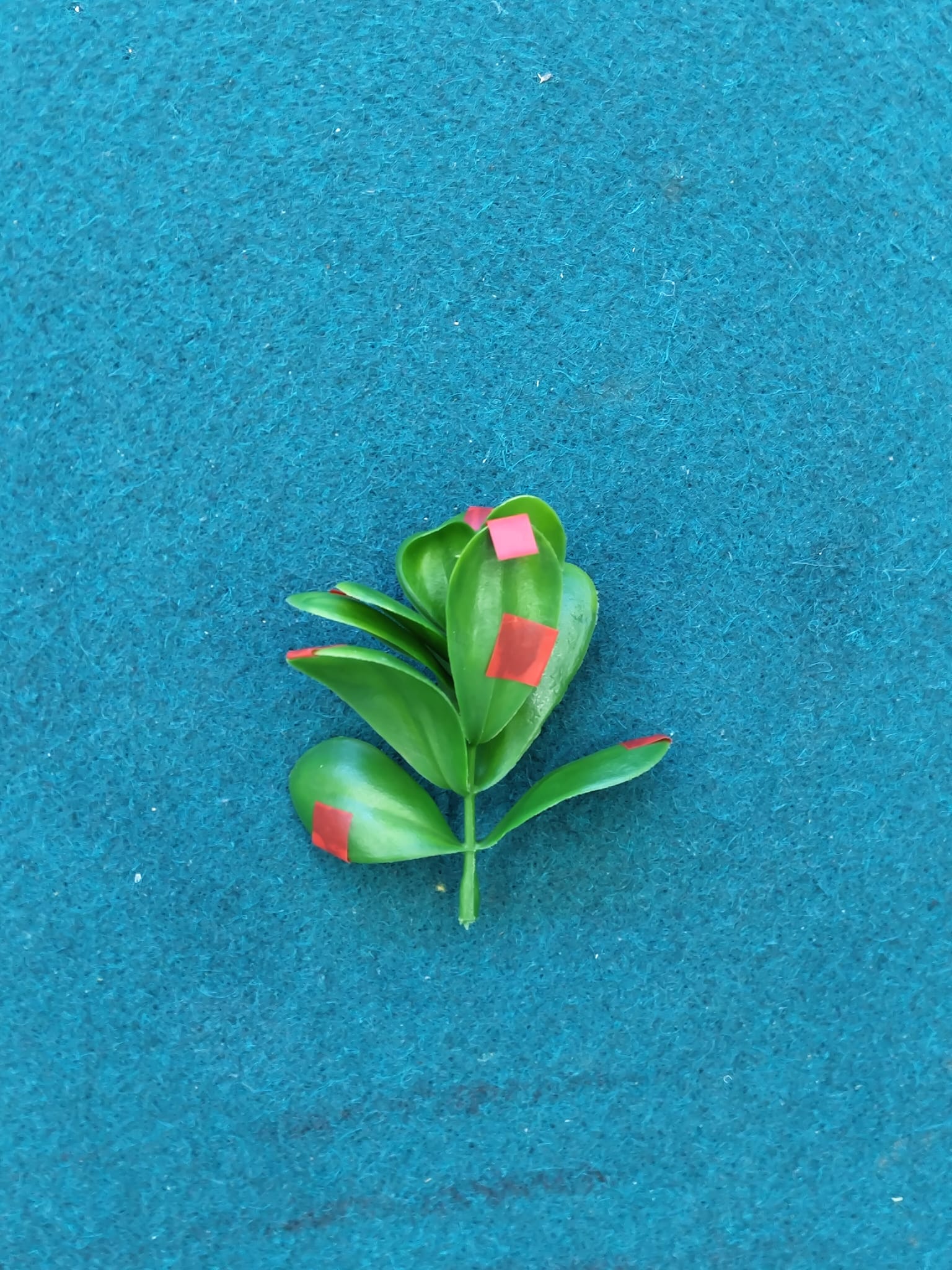}\label{fig:normal-ill}}%
\hfill
\subfloat[Medium healthy crop (\texttt{A})]{\includegraphics[width=.32\linewidth,trim={0cm, 6cm, 0cm, 12cm},clip]{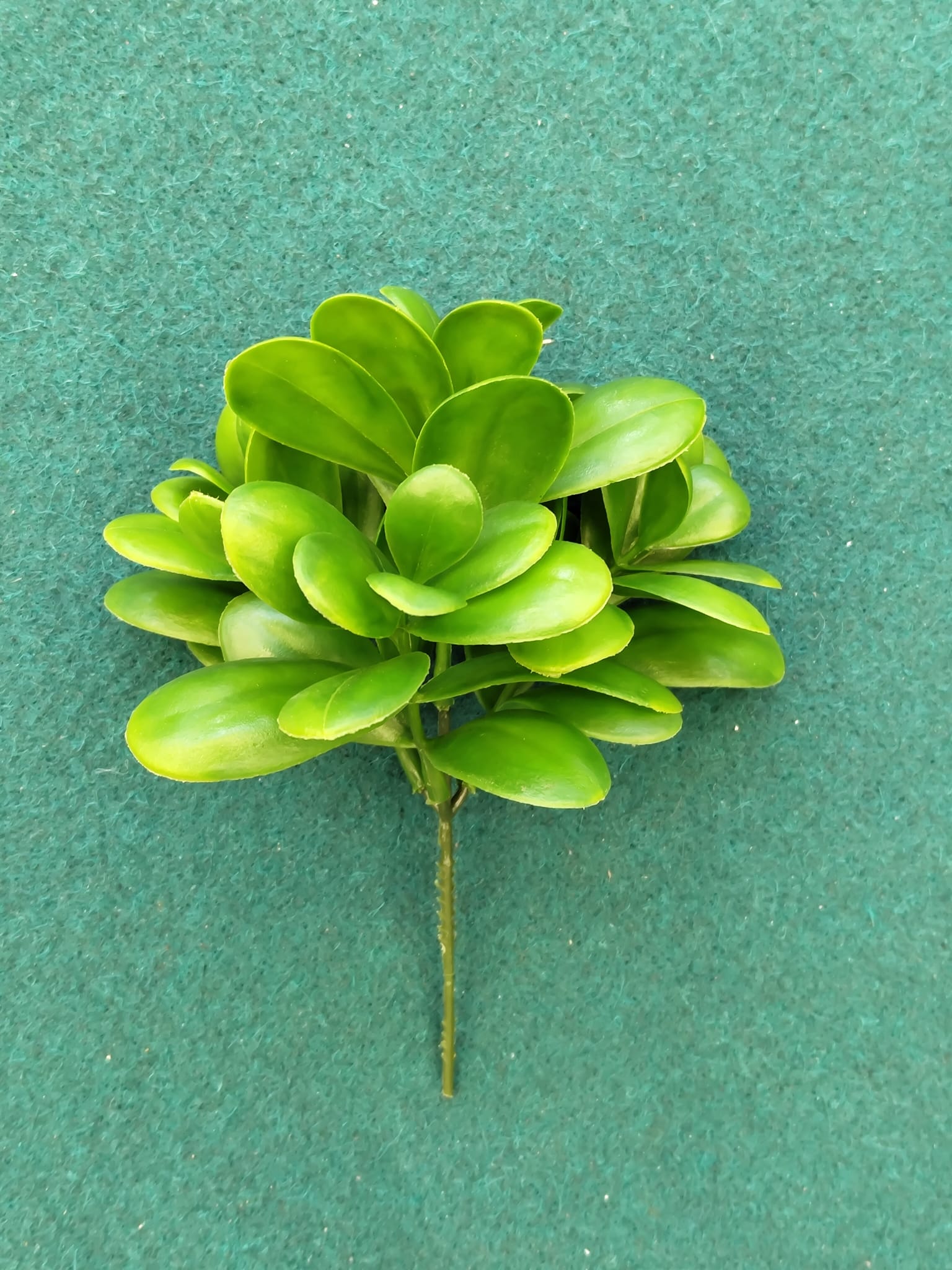}\label{fig:medium-healthy}}%
\hfill
\subfloat[Small healthy weed (\texttt{B})]{\includegraphics[width=.32\linewidth,trim={0cm, 4cm, 0cm, 14cm},clip]{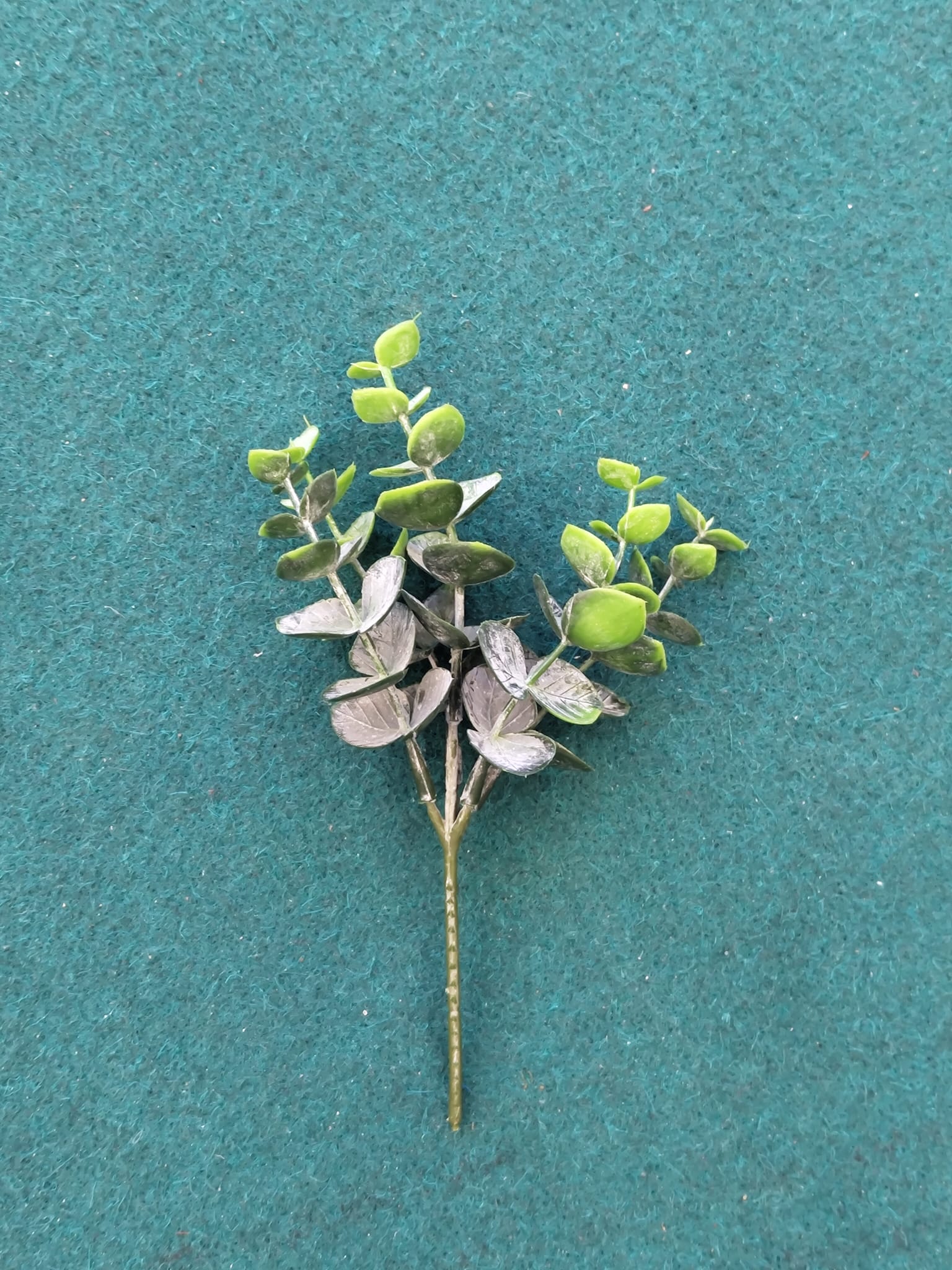}\label{fig:weed-healthy}}
\vspace{2mm}
\caption{Different types of used artificial plants}
\label{fig:plant-types}
\end{figure}

For the experiments, two different types of artificial plants were placed in rows on an indoor (robot soccer) field, see Fig.~\ref{fig:plant-types}. The rows were approximately 1~m apart, and within the rows the plants were approximately 0.3~m apart. Four artificial plants of the same kind were put together to mimic a larger plant size (see Fig.~\ref{fig:medium-healthy}). Pieces of red tape were placed on some of the plants to indicate a disease (see Fig.~\ref{fig:normal-ill}). This simplifies the experiments, as the tapes are a clear visual feature, making them detectable by a YOLOv8n model. As the research of Bhimte and Thool shows, in a real outdoor environment, plant diseases can also be detected using simple vision processing techniques \citep{Bhimte2018DiseaseVision}.

For the detection of these artificial plants, the YOLOv8n model provides image segmentation and object classification. One disadvantage of Spot is that the onboard cameras only provide low-resolution grayscale images, and these can, therefore, not be used as an input for the YOLOv8n model.
Hence the additional RealSense L515 LiDAR camera was mounted on top of Spot. In combination with the depth information obtained by the RealSense camera, a position estimate of the artificial plants can be made. This process is shown in Appendix~\ref{app:object-detection}. For image segmentation, no additional color filter was used, since the indoor field is green as well, and since the camera exposure was changing in different situations (as can be seen in Fig.~\ref{fig:plant-types}), no general color filter parameters could be determined.

For the data association $w_{\mathrm{sem}}=0.3$ was chosen for these experiments as well, since false detections occurred quite regularly.
The number of object measurements after which the row estimation is performed is $n_{\text{det,est}} = 12$ (see Section~\ref{sec:row-information}) measurements.
A summary of the different plant attributes and their possible values used for the experiments is given in Table~\ref{tab:attributes-exp}.

\begin{table}[ht]
    \centering
    \caption{Plant attributes and their possible values used for the experiments}
    \label{tab:attributes-exp}
    \begin{tabular}{l|l|l|l}
        \textbf{Experiment} & \textbf{Plant type} & \textbf{Size} & \textbf{Health status} \\ \hline
        Mixed intercropping & \texttt{A}, \texttt{B} & \texttt{small}, \texttt{medium}, \texttt{big} & \texttt{healthy}, \texttt{diseased} \\ \hline
        Crops and weeds & \texttt{crop}, \texttt{weed} & \texttt{small}, \texttt{medium}, \texttt{big} & \texttt{healthy}, \texttt{diseased}
    \end{tabular}
\end{table}

\subsubsection{Experimental results}
\begin{figure}
    \centering
    \includegraphics[width=\linewidth]{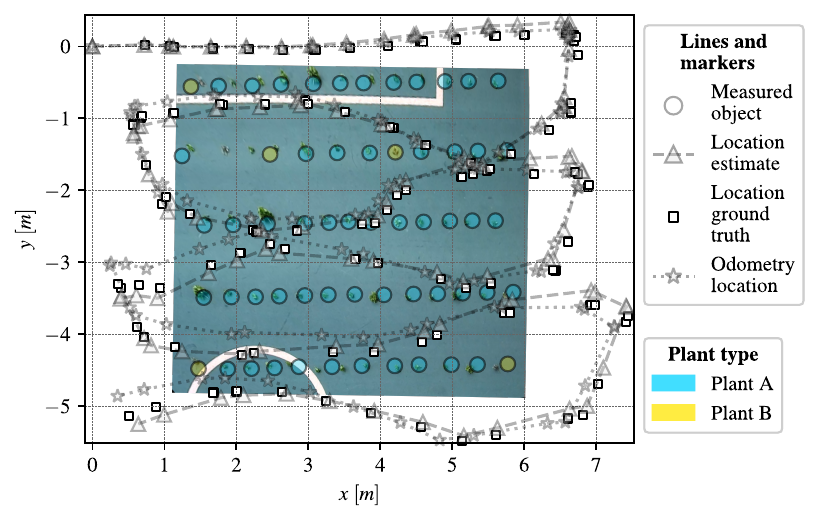}
    \caption{Experimental result, with two different types of plants}
    \label{fig:plants-in-row}
\end{figure}

\paragraph{Emulated mixed intercropped field}
Fig.~\ref{fig:plants-in-row} shows an experimental result for a mixed intercropped field with two different types of plants, \texttt{A} and \texttt{B} (see Fig.~\ref{fig:plant-types}). Here, out of the 57 plants of type \texttt{A}, 54 were detected correctly, and out of the 5 plants of type \texttt{B}, 4 were detected correctly, after only one trip through the field. Furthermore, one plant of type \texttt{A} was incorrectly indicated as a plant of type \texttt{B}, and three plants were not detected. This figure also shows that noise in the position estimate from the odometry is reduced by the graph-based SLAM framework. The ground truth positions were established using an OptiTrack Motion Capture system.
Furthermore, Fig.~\ref{fig:plants-in-row} shows that the merging of two plants is not always successful, as is indicated by the two plants in the bottom row, right to the center circle of the field, around $(3, -4.5)$. Here, two individual plants are represented in the model as one single plant, and new measurements of either of those plants are likely associated with the single plant in the world model. However, this might be corrected if the two plants are detected at the same time in one picture. Using MHT could solve this type of ambiguous situations as well, but comes with computational disadvantages as explained in Section~\ref{sec:sensor-fusion}.

In Appendix \ref{app:diseases}, the results concerning disease detection are shown for the same experiment.

\begin{figure}
\centering
   \includegraphics[width=0.75\linewidth]{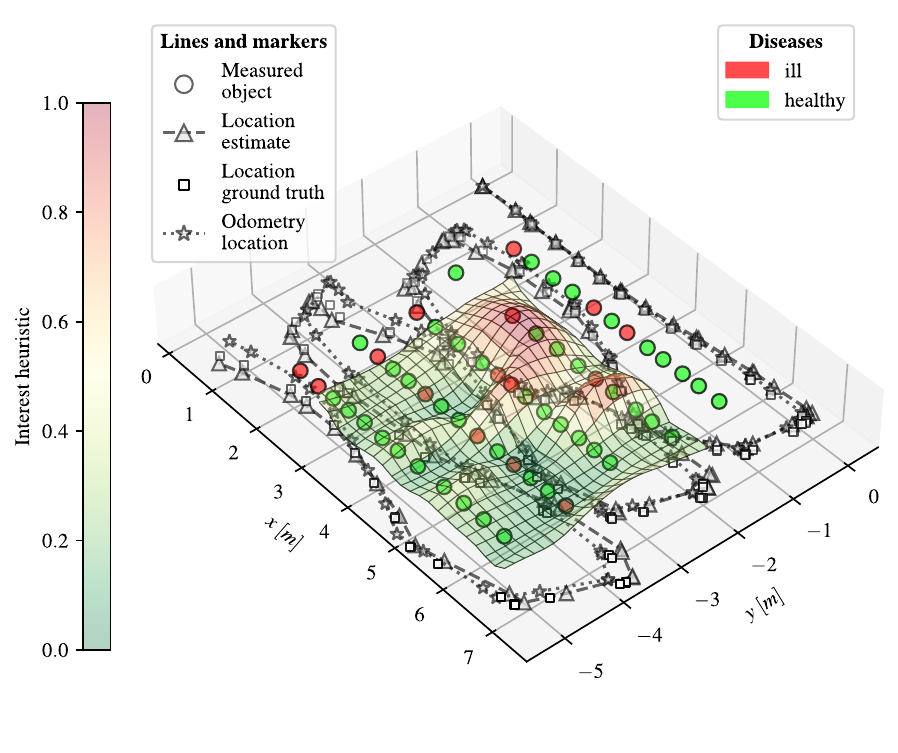}
\vspace{-0.5cm}%
\caption{World model representation, showing the health of the plants, where the colored surface represents the interest heuristic over the field}
\label{fig:interest-heuristic}
\end{figure}

The evaluation of the interest heuristic, introduced in Section~\ref{sec:heuristic}, can be seen in Fig.~\ref{fig:interest-heuristic}. Parts of the field where the world model is uncertain about could for example be visited again, so that additional measurements result in a more certain belief of those plants.

\begin{figure}
    \centering
    \includegraphics[width=0.9\linewidth]{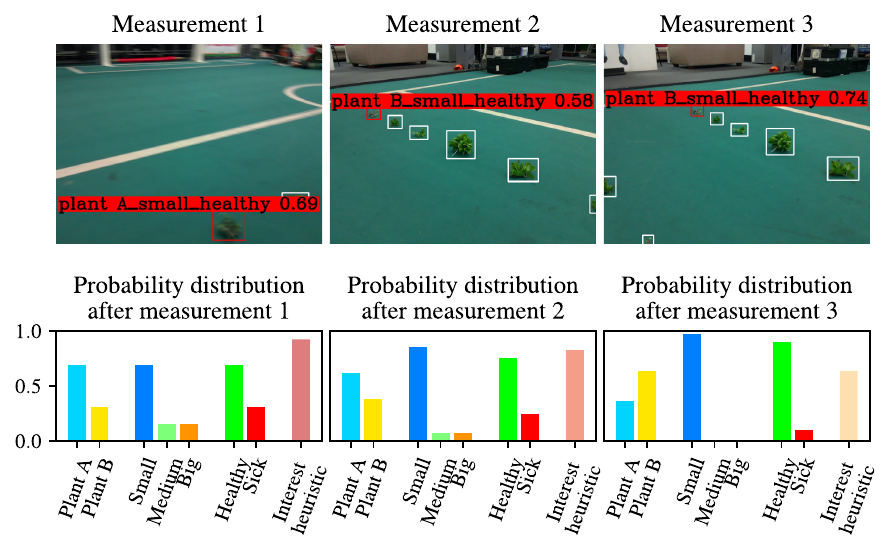}
    \caption{Handling of inconsistent measurements by the proposed framework, where the red detection is the same plant (plant \texttt{B}) in the field}
    \label{fig:vague-measurements}
\end{figure}

The proposed world modeling framework is especially useful in the case of inconsistent or uncertain measurements since different inconsistent or uncertain measurements can still form an appropriate prediction for the measured object.
An example is shown in Fig.~\ref{fig:vague-measurements}, where at three different moments and positions, different measurements of a small healthy weed plant are made.
There is one wrong measurement, falsely indicating that the weed is a plant, and two correct measurements.
For the type of plant, the probability distribution is corrected towards a higher probability for the weed after the three measurements, while still keeping some uncertainty, which is also indicated by the `Interest heuristic' bar.
Since the other two semantic properties are measured correctly all the time, the world model is more certain about these properties.

\paragraph{Crops and weeds}
Furthermore, in Appendix \ref{app:not-in-row}, additional experimental results are shown, where artificial weed plants are considered. In this experiment, the weed plants are not constrained to a row of crops (as would be the case in real life).

\section{Conclusion and discussion}
\label{sec:conclusion}
This final section reflects on the goals and assumptions made. The conclusion elaborates on the achievements that are made concerning the envisioned contributions and the discussion reflects on the assumptions and potential issues of the proposed framework. Additionally, we present suggestions for future work.

\subsection{Conclusion}
This work proposes a probabilistic semantic world modeling framework using Bayesian inference and a $g^2o$ graph-based SLAM method.
This framework is specifically designed for precision agriculture purposes, such as intercropping.
For these precision farming purposes, prior knowledge of an agricultural field, such as the knowledge that plants are planted in rows and the distance between those rows, can be used to boost the performance of the modeling framework (Contribution~\ref{contribution:main}).
The framework does not rely on GPS, but instead, a graph-based SLAM solution was used, with the use of semantic measurements.
Here, plants and their semantic properties are detected using a trained YOLOv8n network, and those semantic properties are saved in a probabilistic manner and updated over time using Bayesian inference in real-time (Contribution~\ref{contribution:inference}). These semantic properties include plant type, plant size, and presence of diseases, as these are relevant for precision agriculture or interesting to the farmer, but other relevant semantic properties, such as ripeness and temperature, can also be included in the world modeling framework.
These semantic properties can be implemented as either discrete or continuous variables.

A semantic- and distance-based GNN data association was used to associate new plant detections with existing plants in the world model.
In addition, a heuristic was developed that indicates how interesting it is to revisit a plant, based on the model's certainty about the its semantic properties and how often they have already been measured (Contribution~\ref{contribution:closedloop}).
Simulations in Gazebo show the performance of the framework, unveiling the usefulness of the $g^2o$ graph-based SLAM. The current model is scalable up to a field with 400 plants for real-time applications, where speed improvements can be made relatively easily to increase this number.

Boston Dynamics robot dog Spot was used to perform experiments on an indoor field with artificial plants.
Although this is not necessarily the perfect robot to use within precision agriculture due to the camera quality and its inability to walk in more difficult agricultural environments, it proved to be an appropriate platform for testing the world modeling framework in a controlled environment. In conclusion, while Spot provides an easy platform for testing the world modeling framework, the world model could be applied to different mobile robots as well.

A general-purpose semantic world modeling framework as provided in this work could be a starting point in standardizing the maintenance of intercropped and precision farming fields. A robot that captures the semantics in such a world model can then be used to perform tasks for the different kinds of crops in one go. Furthermore, the world model can also be visualized to give farmers an oversight of their crop fields, with additional insight into the semantic properties that are of their interest.

\subsection{Discussion}
Possible improvements of the world modeling framework include improving object detections, including plant models, and including dependencies between semantic properties.

A major improvement of the proposed world modeling framework lies in the detection of objects and their semantics.
The used YOLOv8n model was trained on a relatively small dataset and could, therefore, be much more robust by increasing the training dataset, at the cost of time-consuming data annotation.
On top of that, stabilizing the camera that is used for the experiments will help to improve image quality, and thus detection quality.

Regarding the plant models, Functional-Structural Plant Models (FSPM) could be used to update numerous semantics. If no measurements are made over time, this will still result in a representative model of the field, and it could also improve data association. Besides, an FSPM can be used to update the simulation environment, to make it more (visually) realistic. Dependencies between different plants can be used to update the semantic properties even more precisely, alleviating Assumption~\ref{asu:independence}. For example, when one plant has a disease, the surrounding plants might have a higher chance of having a disease as well.

Furthermore, when conducting experiments in a field over multiple days, the localization within the field might be more challenging, since the robot will not have sight of the field all the time, for example if the mobile robot has to be charged.
Within the current framework, it is not possible to drop the robot in the field and let it localize itself using a previously mapped semantic world model.
Furthermore, if multiple robots were used simultaneously, the data association between objects measured by both robots would get more complicated, and making hard decisions would probably not work.
It would be relatively easy to use GPS measurements in the proposed framework to solve this problem, but as stated before, these measurements are not always reliable. Alternatively, as possible future work, a multiple-hypothesis approach can be implemented as a solution to this localization and multiple-robot data association problem.
This might also be useful if there is a higher uncertainty on the odometry data, or when plants are too similar, making the data association more susceptible to errors.
However, since there will be many similar plants within a crop field, a multiple-hypothesis approach is likely to result in a large number of hypotheses, making this approach challenging, especially for real-time application.
Appropriate hypothesis pruning methods will be required to solve this challenge.

Future work also includes more autonomous movement within the field, for example, based on the rows of the field.
Currently, a predefined path is used to let the robot do its first surveillance, but when detecting the rows, this information can also be used to let the robot walk autonomously along the rows. If the robot then detects the end of a row, the robot could automatically turn to the next row.
Next to that, supervisory control based on the developed interest heuristic can be developed, by using time and distance optimization.

Finally, and most importantly, experiments should be performed in a real field of crops. In the end, such experiments would yield the most important insights and improvements for the world modeling framework, due to the difficult practical complications in such an environment, such as the lighting conditions and the ability of the mobile robot to move through the field.
Once experiments in a real field of crops are successful, the proposed semantic world modeling framework can facilitate the development of autonomous robots in precision farming, helping to ensure a sustainable future for agriculture.

\section*{Acknowledgment}
This research was conducted as part of the Synergia consortium.
The Synergia project is organized and led by Wageningen University and Research in 
close cooperation with Next Food Collective as well as the Universities of Delft, Twente, 
Eindhoven, and Nijmegen. The authors have declared that no competing interests exist 
in the writing of this publication. Funding for this research was obtained from the 
Netherlands Organisation for Scientific Research (NWO grant 17626), IMEC-One Planet 
and other private parties.

\bibliographystyle{plainnat}
\bibliography{references}

\begin{appendices}
\section{Object detection}
\label{app:object-detection}
\begin{figure*}
    \centering
    \subfloat[Original RealSense L515 image\label{fig:original}]{\includegraphics[width=0.49\textwidth]{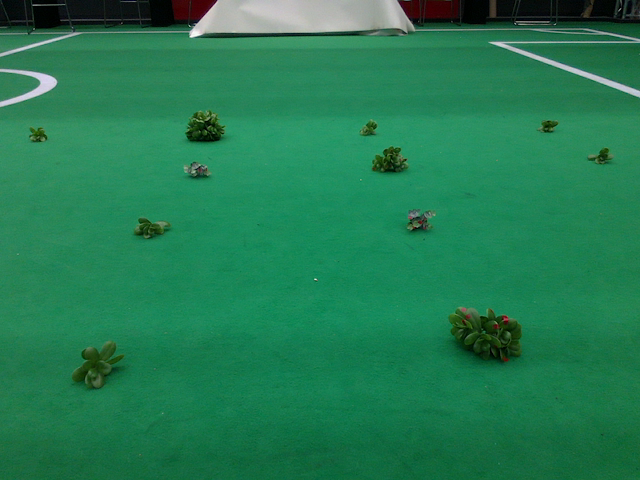}}%
    \hfill
    \subfloat[YOLOv8n detections and segmentation\label{fig:detections}]{\includegraphics[width=0.49\textwidth]{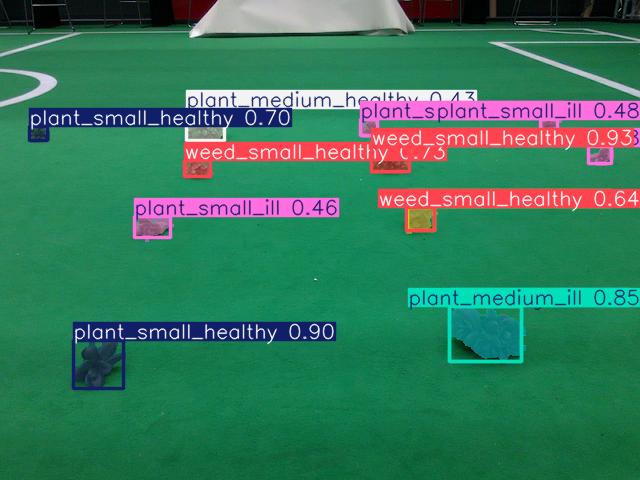}}
    
    \subfloat[RealSense L515 depth visualization\label{fig:depth}]{\includegraphics[width=0.49\textwidth]{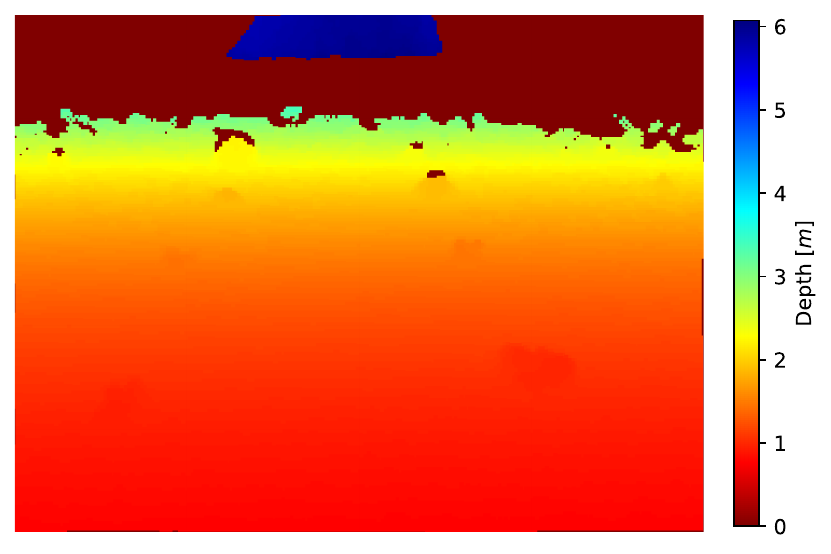}}%
    \hfill
    \subfloat[Detections in the camera view\label{fig:measurement}]{\includegraphics[width=0.465\textwidth]{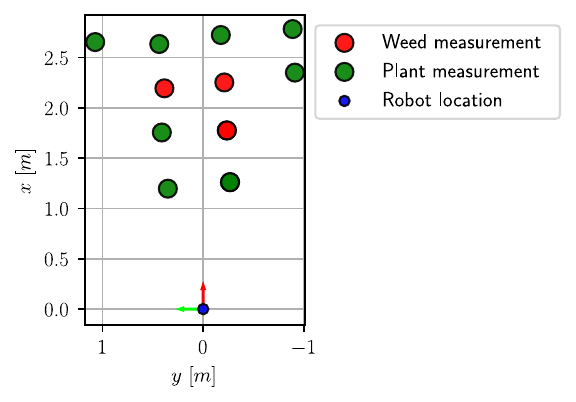}}

    \vspace{2mm}

    \caption{Steps of the object detection using YOLOv8n and a RealSense camera}
    \label{fig:object-detection}
\end{figure*}

For the object detection in the Gazebo simulation, automatic annotations were generated based on the positions of the plants with respect to the simulated camera in the Gazebo environment. This allowed us to get a large annotated data set of 1741 pictures. The trained YOLOv8n model using this dataset had a mean Average Precision (mAP) of 0.711 at an Intersection over Union (IoU) of 50, and an average mAP of 0.579 for IoUs between 50 and 95, indicating that the model still had difficulty pinpointing objects accurately, which could lead to wrong position estimations of the Gazebo plants. The precision of the trained model was 0.834, which is high enough to handle objects correctly within the proposed framework.
However, the training of the YOLOv8n network for a more realistic situation is much more relevant, even though artificial plants are used.
For the object detection of the artificial plants on the indoor field, a small set of 220 photos of a field in which the artificial plants were laid down was used to manually annotate the different types of artificial plants in the pictures. After training a YOLOv8n model using this dataset, a somewhat larger set of 411 photos was annotated and segmented using the previously trained model, combined with a Segment Anything Model (SAM) \citep{Kirillov2023SAM}. This dataset was manually verified, after which a final YOLOv8n model with image segmentation was trained. An exemplary result of this model is shown in Fig.~\ref{fig:original} and Fig.~\ref{fig:detections}. This trained model had a mAP of 0.954 at an IoU of 50, and an average mAP of 0.819 for IoUs ranging between 0.50 and 0.95, indicating that the bounding boxes are pinpointed better than for the Gazebo plants. Furthermore, the trained model has a precision of 0.910 in the trained dataset.

The used artificial plants consist of two types of plants (\texttt{plant} and \texttt{weed}), three different sizes (\texttt{small}, \texttt{medium}, and \texttt{big}), and plants with a binary health status (\texttt{diseased} or \texttt{healthy}).
Rather than training the model with three individual tags per segmented object (for instance, \texttt{plant}, \texttt{small} and \texttt{healthy}), the YOLOv8n model was trained with tags containing all semantic properties (for instance, `plant\_small\_healthy'), as this provided better results.
The reason for this is probably that during training, the model cannot handle multiple identical bounding boxes with different tags, as it will mark training detections as false positives, while those detections are true positives.
It might be better to use different YOLO models for each semantic property, or use one YOLO model for detecting individual plants, and perform image classification on each bounding box.
However, this would take significantly more time, as more images and more models have to be run per detection.
When more than three semantic properties have to be detected, it might still be better to use image classification of the bounding boxes of individual plants.
Although the type of plant and the health status of plants are detected correctly most of the time, the current model has more difficulty detecting the size of the plants.
This, however, can be used to our benefit as this uncertainty in detections can be used to show how the framework handles this in a probabilistic manner.

Next, with the use of the depth information by the L515 RealSense camera, from which a depth visualization is shown in Fig.~\ref{fig:depth}, a position estimate was made using the intrinsic camera matrix and camera position with respect to the body-fixed frame of the robot. These position estimates can be seen in Fig.~\ref{fig:measurement}. Here, it must be noted that objects with a distance larger than 2.5~m from the camera were not considered for higher performance.

\section{Discrete time step update}
\label{app:time-update}
This appendix explains how a discrete probability distribution for a (size) variable can be updated over time, using a time-dependent growth model.

We first consider a continuous growth model for the (continuous) plant length $x$:
\[
    \dot{x} = \alpha \cdot (l_\mathrm{max} - x),
\]
where $\alpha$ is a growth factor, and $l_\mathrm{max}$ is the maximum plant length. This model can be discretized regarding time, resulting in
\[ \begin{aligned}
    x(t+\Delta t) &= f(x(t), \Delta t) \\
    &= l_\mathrm{max} - (l_\mathrm{max} - x(t))\cdot e^{-\alpha \Delta t}.
\end{aligned} \]
If we now also discretize the plant length into $n$ steps, we can represent $x$ as a vector $\xi$ with the probability distribution for the discretized segments of the plant length.
This probability distribution vector can be updated by using a Bayesian inference step after a measurement, but also over time, based on the plant model given above, leading to
$$\xi(t+\Delta t) = A(\Delta t) \cdot \xi(t).$$
Each entry $i,j$, denoted by $a_{ij}$ of matrix $A(\Delta t)$ coincides with the probability that the state is in cell $j$ at time $t+\Delta t$ when uniformly started at cell $i$ at time $t$. These are given by
\[
    a_{ij} = \left\{ \begin{array}{ccc}
        0 &\mathrm{ for} &~p_{ij} \leq 0 \\
        p_{ij} - \sum_{k=1}^{i-1} a_{kj} &\mathrm{ for} &~0 < p_{ij} < 1 \\
        1 &\mathrm{ for} &~p_{ij} \geq 1.
    \end{array} \right.
\]
where $p_{ij}$ is given by
\[
    p_{ij} = \dfrac{f^{-1}\left(\dfrac{i}{n}l_\mathrm{max}, \Delta t\right)-\dfrac{j-1}{n}l_\mathrm{max}}{\dfrac{l_\mathrm{max}}{n}}.
\]

\begin{figure}
    \centering
    \includegraphics[width=0.8\linewidth]{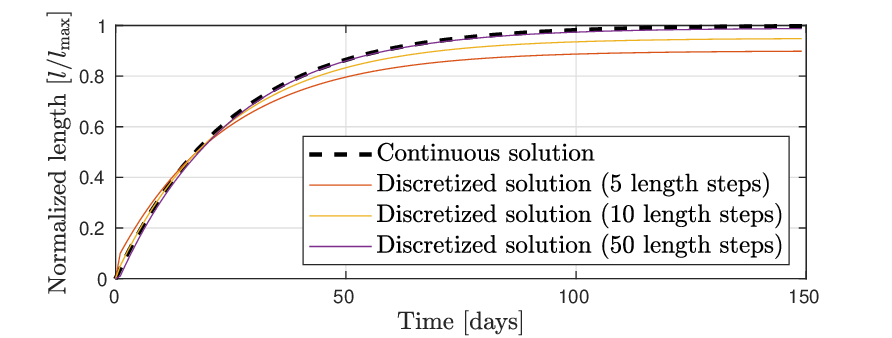}
    \caption{Weighted mean of a discretized probability distribution for length over time, with ${\Delta t = 1~[\mathrm{days}]}$ and ${\alpha = 0.04}$, for different length discretization steps.}
    \label{fig:discretized-time}
\end{figure}

In Fig.~\ref{fig:discretized-time}, the time evolution of the weighted mean of a discretized probability distribution for different discretization steps is shown.

Stochastic model disturbances can be added with a similar procedure that is omitted for the sake of brevity.

\section{Additional simulation results}
\label{app:simulations}
\begin{figure}
    \centering
    \includegraphics[width=0.8\linewidth]{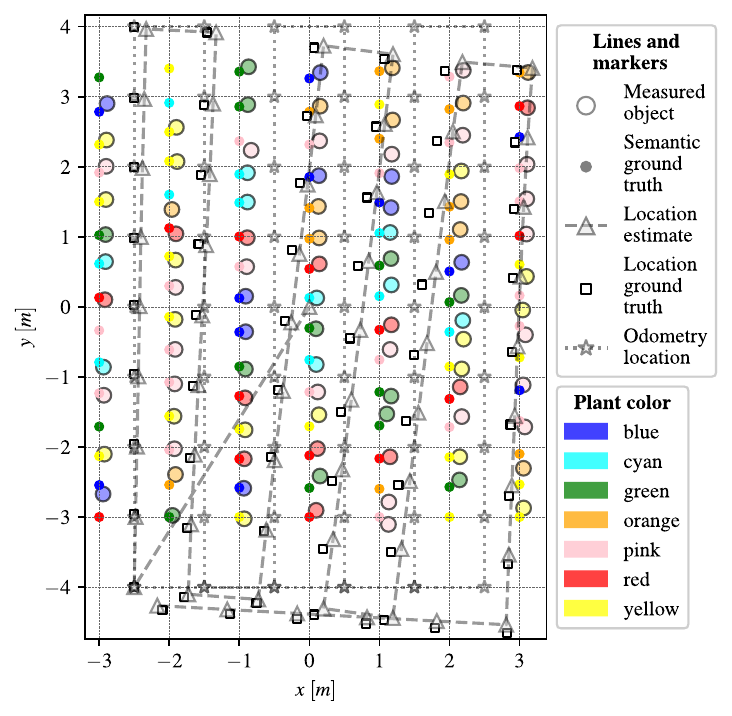}
    \caption{Odometry noise visualization after returning to the start of the traveled path, where the error without graph optimization is 0.516~m, and the error with graph optimization is 0.093~m.}
    \label{fig:app-odom-noise-after}
\end{figure}

The simulation results in Fig.~\ref{fig:odom-noise} (from Section~\ref{sec:simulations}) and Fig.~\ref{fig:app-odom-noise-after} show how the odometry noise can remain limited by returning to a point that has been visited before, in this case, the start of the traveled path. In the simulation of Fig.~\ref{fig:app-odom-noise-after}, the plants that were initiated at the start of the traveled path were correctly associated when measured again at the end of the surveillance trip. This will ensure that the created map is consistent. For that purpose, the localization error should not increase too much, which can be achieved by returning to existing plants in the world map every now and then. Alternatively, a unique landmark (such as a fiducial marker) could be used in the future to provide consistency.

\section{Additional experimental results}
\label{app:experiments}
This appendix shows additional experimental results using the proposed world modeling framework.
\subsection{Experiment with disease detection}
\label{app:diseases}
\begin{figure}
    \centering
    \includegraphics[width=0.9\linewidth]{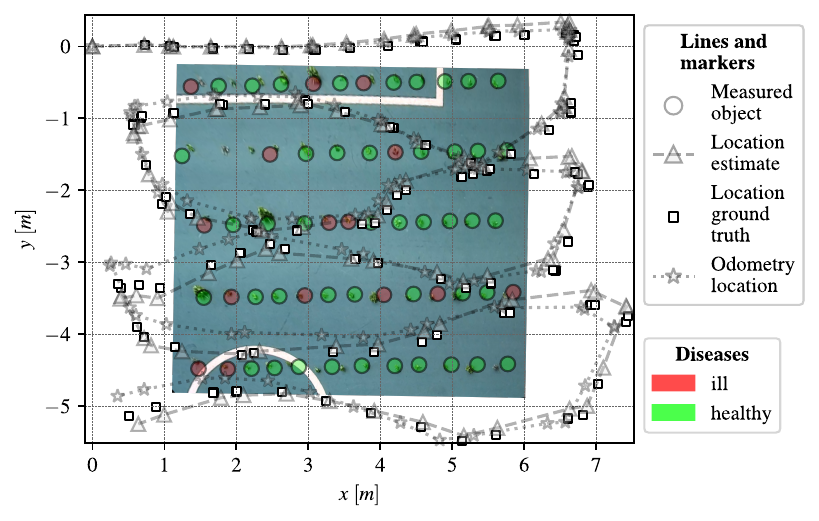}
    \caption{Results of the experiment in which diseases are detected}
    \label{fig:diseases}
\end{figure}

For the experimental result in Fig.~\ref{fig:diseases}, the detected plants are shown, where the color indicates whether the plant is believed to be healthy or diseased. Of the 18 plants with red tape, indicating a disease, 14 plants were correctly perceived as diseased. Of the 44 plants without red tape (healthy), 40 plants were correctly indicated as healthy. Five plants were detected with an incorrect health status and three plants were not detected at all.

\subsection{Experiment with weeds not constrained to rows}
\label{app:not-in-row}
\begin{figure}
    \centering
    \includegraphics[width=0.9\linewidth]{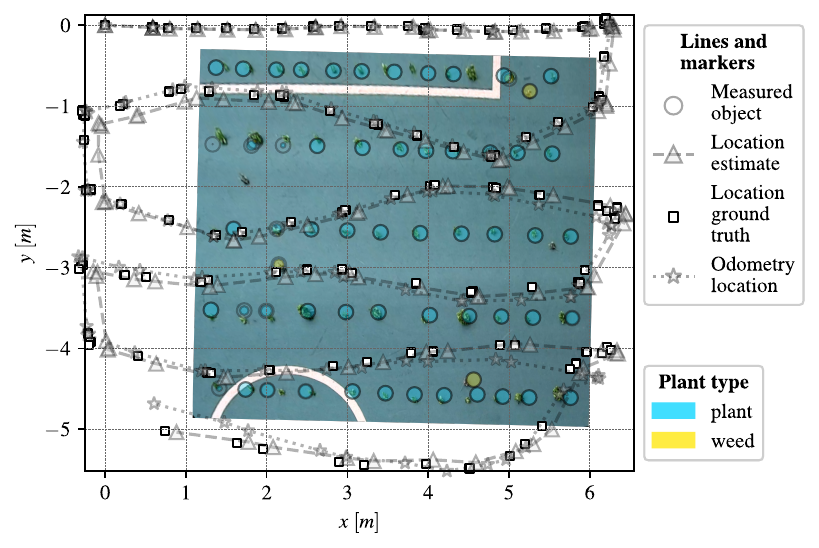}
    \caption{Results of the experiment in which weed plants (yellow) are not constrained to rows}
    \label{fig:not-in-row}
\end{figure}

In Fig.~\ref{fig:not-in-row}, an experiment is shown in which the weeds are not constrained to the rows, while the crops do have soft constraints to the rows. Of the five weeds in the field, three are detected correctly. One of the undetected weeds (between the second and third row, at the left) was not detected by the YOLOv8n model and also the other weed (third row, at the right) was not detected, possibly because it is too close to a crop.

\end{appendices}

\end{document}